\documentclass{article}

\PassOptionsToPackage{numbers, compress}{natbib}

\usepackage[preprint]{neurips_2021}

\usepackage[utf8]{inputenc} \usepackage[T1]{fontenc}    \usepackage{hyperref}       \usepackage{url}            \usepackage{booktabs}       \usepackage{amsfonts}       \usepackage{nicefrac}       \usepackage{microtype}      \usepackage{xcolor}         

\usepackage{hyperref}
\usepackage{url}
\usepackage{xcolor}
\usepackage{amsthm}
\usepackage{amsmath}
\usepackage{amssymb}
\usepackage{graphicx} 
\usepackage{booktabs} 
\usepackage{float}
\usepackage{bbold}
\usepackage{notoccite}

\newcommand{\bR}{\mathbb{R}}
\newcommand{\bE}{\mathbb{E}}

\newcommand{\din}{d}
\newcommand{\dout}{D}

\newcommand{\duality}{Parameter-Space / Function-Space }
\newcommand{\wick}{\text{Wick}}

\renewcommand{\iota}{i}

\newtheorem{theorem}{Theorem}

\newtheorem{corollary}{Corollary}[theorem]

\newcommand{\pthetag}{\mathcal{P}(\theta_g)}

\title{Symmetry-via-Duality: \vspace{.1cm} \\ Invariant Neural Network Densities \\ from Parameter-Space Correlators}

\author{  Anindita Maiti, Keegan Stoner, and James Halverson \thanks{Equal contributions by Maiti and Stoner.}  \vspace{.3cm} \\
  \vspace{.3cm}
  The NSF AI Institute for Artificial Intelligence and Fundamental Interactions \\
  Department of Physics\\
  Northeastern University\\
  Boston, MA 02115 \vspace{.3cm} \\ 
    \texttt{\{maiti.a, stoner.ke, j.halverson\}@northeastern.edu} \\
                                          }

\begin{document}

\maketitle

\begin{abstract}
  Parameter-space and function-space provide two different duality frames in which to study
  neural networks. We demonstrate that symmetries of network densities may be determined via dual computations of network correlation functions, even when the density is unknown and the network is not equivariant. Symmetry-via-duality relies on invariance properties of the correlation functions, which stem from the choice of network parameter distributions. Input and output symmetries of neural network densities are determined, which recover known Gaussian process results in the infinite width limit. The mechanism may also be utilized to determine symmetries during training, when parameters are correlated, as well as symmetries of the Neural Tangent Kernel. We demonstrate that the amount of symmetry in the initialization density affects the accuracy of networks trained on Fashion-MNIST, and that symmetry breaking helps only when it is in the direction of ground truth.
\end{abstract}

\section{Introduction}

Many systems in Nature, mathematics, and deep learning are described by densities over functions. In physics, it is central in quantum field theory (QFT) via the Feynman path integral, whereas in deep learning it explicitly arises via a correspondence between infinite networks and Gaussian processes. 

More broadly, the density associated to a network architecture is itself of foundational importance. Though only a small collection of networks is trained in practice, due to computational limitations, a priori there is no reason to prefer one randomly initialized network over another (of the same architecture). In that case, ideally one would control the flow of the initialization density to the trained density, compute the trained mean $\mu(x)$, and use it to make predictions. Remarkably, $\mu(x)$ may be analytically computed for infinite networks trained via gradient flow or Bayesian inference \cite{Jacot2018NeuralTK,neal,Lee2019WideNN}. 

In systems governed by densities over functions, observables  are strongly constrained by symmetry, which is usually determined via experiments. Examples include the Standard Model of Particle Physics, which has gauge symmetry $SU(3)\times SU(2) \times U(1)$ (possibly with a discrete quotient), as well as certain multi-dimensional Gaussian Processes. In the absence of good experimental data or an explicit form for the density, it seems difficult 
to deduce much about its symmetries.

We introduce a mechanism for determining the symmetries of a neural network density via duality, even for an unknown density. A physical system is said to exhibit a duality when it admits two different, but equally fundamental, descriptions, called duality frames. Hallmarks of duality include the utility of one frame in understanding a feature of the system that is difficult to understand in the other, as well as limits of the system where one description is more tractable than the other. In neural networks, one sharp duality is \duality duality: networks may be thought of as instantiations of a network architecture with fixed parameter densities, or alternatively as draws from a function space density. In GP limits where a discrete hyperparameter $N \to \infty$ (e.g. the width), the number of parameters is infinite and the parameter description unwieldy, but the function space density is Gaussian and therefore tractable. Conversely, when $N = 1$, the function space density is generally non-perturbative due to large non-Gaussianities, yet the network has few parameters.

We demonstrate that symmetries of network densities may be determined via the invariance of correlation functions computed in parameter space. We call this mechanism symmetry-via-duality, and it is utilized to demonstrate numerous cases in which transformations of neural networks (or layers) at input or output leave the correlation functions invariant, implying the invariance of the functional density. It also has implications for learning, which we test experimentally. For a summary of our contributions and results, see Section \eqref{sec:discussion}.

\paragraph{Symmetries, Equivariance, and Invariant Generative Models Densities.} 
Symmetries of neural networks are a major topic of study in recent years. Generalizing beyond mere invariance of networks, equivariant networks \cite{cohen2018intertwiners,mohamed2020data,falorsi2018explorations, cohen2016group, weiler20183d, fuchs2020se3transformers, maron2019invariant, maron2020learning, maron2019universality, thomas2018tensor, ravanbakhsh2017equivariance, kondor2018clebschgordan, higgins2018definition, kondor2018generalization, anderson2019cormorant, bogatskiy2020lorentz, worrall2017harmonic} have aided learning in a variety of contexts, including gauge-equivariant networks \cite{gaugeCNNscohenwelling,bsplineCNNs} and their their utilization in generative models \cite{DBLP:journals/corr/abs-1902-01967, DBLP:journals/corr/abs-1909-02775,rezende2019equivariant,kohlerkleinnoe}, for instance in applications to  Lattice QCD \cite{Kanwar_2020,Boyda:2020hsi}. See also \cite{Betzler:2020rfg,Krippendorf:2020gny} for symmetries and duality in ML and physics.

Of closest relation to our work is the appearance of symmetries in generative models, where invariance of a generative model density is often desired. It may be achieved via draws from a simple symmetric input density $\rho$ on $V$ and an equivariant network $f_\theta: V \to V$, which ensures that the induced output density $\rho_{f_\theta}$ is invariant. In Lattice QCD applications, this is used to ensure that gauge fields are sampled from the correct $G$-invariant density $\rho_{f_\theta}$, due to the $G$-equivariance of a trained network $f_\theta$.

In contrast, in our work it is the network $f_\theta$ itself that is sampled from an invariant density over functions. That is, if one were to cast our work into a lattice field theory context, it is the networks themselves that are the fields, and symmetry arises from symmetries in the density over networks. Notably, nowhere in our paper do we utilize equivariance.

\paragraph{Modeling Densities for Non-Gaussian Processes.} One motivation for understanding symmetries of network densities is that it constrains the modeling of neural network non-Gaussian process densities using techniques from QFT \cite{Halverson_2021} (see also \cite{Cohen_2021}), as well as exact non-Gaussian network priors \cite{zavatoneveth2021exact} on individual inputs. Such finite-$N$ densities arise for network architectures admitting a GP limit \cite{neal, williams, lee, cohen2020learning,Matthews2018GaussianPB,yangTP1,yangTPorig,Novak2018BayesianCN,GarrigaAlonso2019DeepCN} as $N\to \infty$, and they admit a perturbative description when $N$ is large-but-finite. These functional symmetry considerations should also place constraints on NTK scaling laws \cite{Dyer2020AsymptoticsOW} and preactivation distribution flows \cite{Yaida2019NonGaussianPA} studied in parameter space for large-but-finite $N$ networks.

\section{Symmetry Invariant Densities via Duality \label{sec:symmetryviaduality}}

Consider a neural network $f_\theta$ with continuous learnable parameters $\theta$.
The architecture of $f_\theta$ and parameter density $P_\theta$ induce a density over functions $P_f$. Let $Z_\theta$
and $Z_f$ be the associated partition functions.
Expectation values may be computed using either $P_\theta$ or $P_f$, denoted as $\bE_\theta$ and $\bE_f$, respectively, or
simply just $\bE$ when we wish to be agnostic as to the computation technique. 

The $n$-point correlation functions (or correlators) of neural network outputs are then 
\begin{equation}
G^{(n)}(x_1,\dots,x_n) = \bE[f(x_1)\dots f(x_n)],
\end{equation}
and (if the corresponding densities are known) they may be computed in either parameter- or function-space.
These functions are the moments of the density over functions. When the output dimension $\dout > 1$, we may write output indices explicitly, e.g. $f_i(x)$, in which case the correlators are written $G^{(n)}_{i_1,\dots,i_n}(x_1,\dots,x_n)$.

Neural network symmetries are a focus of this work. Consider a continuous transformation
\begin{equation}
f'(x) =\Phi(f(x')),
\end{equation}
i.e. the transformed network $f'$ at $x$ is a function $\Phi$ of the old network $f$ at $x'$. We say there is a classical symmetry if $P_f$ is invariant under the transformation,
which in physics is usually phrased in terms of the action $S_f = -{\rm log}\, P_f$. If the functional measure $\textit{Df}\,$ is also invariant, it is said that there is a quantum symmetry and the correlation functions are constrained by 
\begin{align}
G^{(n)}&(x_1,\dots,x_n) = \bE[f(x_1)\dots f(x_n)] \nonumber \\ 
 &= \bE[\Phi(f(x_1'))\dots \Phi(f(x_n'))] =: G'^{(n)}(x_1',\dots,x_n').
 \label{eqn:correlatorconstraint}
\end{align}
See appendix for the elementary proof. In physics, if $x=x'$ but $\Phi$ is non-trivial the symmetry is called internal, and if $\Phi$ is trivial but $x\neq x'$ it is called a spacetime symmetry. Instead, we will call them output and input symmetries to identify the part of the neural network that is transformed; examples include rotations of outputs and translations of inputs. Of course, if $f_{\theta}$ is composed with other functions to form a larger neural network, then input and output refer to those of the layer $f$.

Our goal in this paper is to determine symmetries of network densities via the constraint \ref{eqn:correlatorconstraint}. For a discussion of functional densities, see Appendix \ref{app:densities}.

\paragraph{A Glimpse of Symmetry from Gaussian Processes.} Further study in this direction is motivated by first deriving a result for one of the simplest function-space densities: a Gaussian process. 

Consider a neural network Gaussian Process (NNGP): a real-valued neural network $f_{\theta,N}$ where $N$ is a discrete hyperparameter such that in the asymptotic limit $N\to \infty$, $f_{\theta,\infty}$ is drawn from a Gaussian process. The simplest example is \cite{neal} a fully-connected single-layer network of width $N$. Suppressing $\theta,N$ subscripts and assuming the $N\to \infty$ limit, a NNGP $f$ can be stacked to obtain a vector-valued neural network $f_i: \bR^\din \to \bR^\dout$. The associated two-point function $G^{(2)}_{i_1i_2}(x_1,x_2) = \delta_{i_1i_2} K(x_1,x_2)$, $K(x_1,x_2)$ is the NNGP kernel ($2$-pt function) associated to $f$; i.e. stacking adds tensor structure to the kernel in the form of a Kronecker delta.

If the NNGP has zero mean, then $G^{(2n+1)}(x_1,\dots,x_n)=0$ for all $n$ and the even-point functions may be computed in terms of the kernel via Wick's theorem,
\begin{align}
  \label{eqn:GPevenpoint}
  G^{(2n)}_{i_1,\dots,i_{2n}}(x_1, \dots, x_{2n}) =  
  \sum_{P \in \wick(2n)} 
  \delta_{i_{a_1}i_{b_1}}\dots \delta_{i_{a_n}i_{b_n}}
  K(x_{a_1},x_{b_1})\dots K(x_{a_n},x_{b_n}) 
\end{align}
where the Wick contractions are defined by the set 
\begin{align}
  \wick(n) = \{P \in \text{Partitions}(1,\dots,n)\,\, | \,\, |p| = 2 \,\, \forall p \in P\}.
\end{align}
We write $P \in \wick(2n)$ as $P=\{(a_1,b_1),\dots,(a_{n},b_{n})\}$. A network transformation $f_i \mapsto R_{ij} f_j$ induces an action on each index of each Kronecker delta in \eqref{eqn:GPevenpoint}. For instance,  as 
$\delta_{ik} \mapsto R_{ij}R_{kl} \delta_{jl} = (R\, R^T)_{ik}= \delta_{ik}$ where the last equality holds for $R\in SO(\dout)$. By this phenomenon, the even-point
correlation functions \eqref{eqn:GPevenpoint} are $SO(\dout)$ invariant. Conversely, if the NNGP has a mean $\mu(x) = G_{i_1}^{1}(x_1) \neq 0$, it transforms with a single $R$ and is not invariant. From the NNGP correlation functions, the GP density has $SO(\dout)$ symmetry iff it has zero mean. This is not surprising, and could be shown directly by inverting the kernel to get the GP density and then checking its symmetry.

However, we see correlation functions contain symmetry information, which becomes particularly powerful when the correlation functions are known, but the network density is not.

\paragraph{\duality Duality.} 
To determine the symmetries of an unknown network density via correlation functions, we need a way to compute them. For this, we utilize duality.

A physical system is said to exhibit a duality if there are two different descriptions of the system, often with different degrees of freedom, that exhibit the same predictions either exactly (an exact duality) or in a limit, e.g., at long distances (an infrared duality); see \cite{Polchinski:2014mva} for a review.
Duality is useful precisely when one perspective, a.k.a. a duality frame, allows you to determine something about the system that would be difficult from the other perspective. Examples in physics include electric-magnetic duality, which in some cases allows a strongly interacting theory of electrons to be reinterpreted in terms of a weakly coupled theory of monopoles \cite{Seiberg:1994rs}, and gauge-gravity duality \cite{Maldacena:1997re}, which relates gravitational and non-gravitational quantum theories via the holographic principle. 

In the context of neural networks, the relevant duality frames are provided by parameter-space and function-space, yielding a \duality duality. In the parameter frame, a neural network is considered to be compositions of functions which themselves have parameters drawn from $P_\theta$, whereas in the function frame, the neural network is considered as an entire function drawn from a function-space density $P_f$. Of course, the choice of network architecture and densities $P_\theta$ determine $P_f$, but they do not appear explicitly in it, giving two different descriptions of the system.

\paragraph{Symmetry-via-Duality.}
Our central point is that symmetries of function-space densities may be determined from correlation functions computed in the parameter-space description, even if the function space density is not known.
That is, it is possible to check \eqref{eqn:correlatorconstraint} via correlators computed in parameter space; if so, then the product $\textit{Df} \, P_f$ is 
invariant. Barring an appearance of the Green-Schwarz mechanism \cite{Green:1984sg} in neural networks, by which $\textit{Df} \, P_f$ is invariant but $\textit{Df}\,$ and $P_f$ are not, this implies that $P_f$ is invariant. This leads to our main result.

\begin{theorem}\label{thm:Sfinvariance}
    Consider a neural network or layer 
    \begin{equation}
        f_\theta: \bR^\din \to \bR^\dout
    \end{equation}
    with associated function space measure $\textit{Df}\,$ and density $P_f$,
    as well as a transformation $f'(x) =\Phi(f(x'))$ satisfying 
    \begin{equation}
        \bE_\theta[f(x_1)\dots f(x_n)] 
    = \bE_\theta[\Phi(f(x_1'))\dots \Phi(f(x_n'))].
    \label{eqn:paramcorrelatorconstraint}
    \end{equation}
    Then $\textit{Df}\, P_f$ is invariant, and $P_f$ is 
    itself invariant if a Green-Schwarz mechanism is not effective.
\end{theorem}
The proof of the theorem follows from the proof of \eqref{eqn:correlatorconstraint} in the Appendix and the fact that correlators may also be computed in parameter space. Additionally, there may be multiple such transformations that generate 
a group $G$ of invariant transformations, in which case $P_f$ is $G$-invariant.

The schematic for each calculation is to transform the correlators by transforming some part of the network, such as the input or output, absorb the transformation into a transformation of parameters $\theta_T \subset \theta$ (which could be all $\theta$), and then show invariance of the correlation functions via invariance of $P_{\theta_T}$. Thus,
\begin{corollary} \label{cor} Symmetries of $P_f$ derived via duality rely on symmetry properties of $P_{\theta_T}$. \end{corollary}

In what follows we will  show that \eqref{eqn:paramcorrelatorconstraint} holds in 
numerous well-studied neural networks for a variety of transformations, without requiring equivariance 
of the neural network. Throughout, we use $Z_\theta$ to denote the parameter space partition function of all parameters $\theta$ of the network.

\paragraph{Example: $SO(\dout)$ Output Symmetry.} We now demonstrate in detail that a linear output layer  leads to $SO(D)$ invariant network densities provided that its weight and bias distributions are invariant. The network is defined by $f_i(x) = W_{ij}g_{j} (x)+ b_i$ where $i=1,\dots,\dout$ and $g_{j}$ is an $N$-dimension postactivation with parameters $\theta_g$. Consider an invertible matrix transformation $R$ acting as as $f_i \mapsto R_{ij} f_j$; we use Einstein summation convention here and throughout. The transformed correlation functions are
\begin{align} \label{eq-SO-output}
  &G'^{(n)}_{i_1\dots i_n} (x_1',\dots,x_n') = \mathbb{E}[R_{i_1j_1}f_{j_1}(x_1)\dots R_{i_nj_n}f_{j_n}(x_n)] \nonumber \\
  &= \frac{1}{Z_\theta} \int DW Db\,  D\theta_g \, R_{i_1j_1}(W_{j_1k_1}g_{k_1}(x_1) + b_{j_1})\dots  R_{i_nj_n}(W_{j_nk_n}g_{k_n}(x_n) + b_{j_n}) P_W P_b P_{\theta_g} \nonumber \\
  &= \frac{1}{Z_\theta} \int |R^{-1}|^2  D\tilde W D\tilde b\,  D\theta_g \, (\tilde W_{i_1k_1}g_{k_1}(x_1) + \tilde b_{i_1})\dots  (\tilde W_{i_nk_n}g_{k_n}(x_n) + \tilde b_{i_n}) P_{R^{-1} \tilde W} P_{R^{-1}\tilde b} P_{\theta_g} \nonumber \\
  &= \mathbb{E}[f_{i_1}(x_1)\dots f_{i_n}(x_n)] = G^{(n)} (x_1,\dots,x_n),
\end{align}
where, e.g., $DW$ denotes the standard measure for all $W$-parameters in the last layer, and the crucial second-to-last equality holds when $|R|=1$, $\smash{P_W =P_{R^{-1}\tilde W}=P_{\tilde W}}$, and $\smash{P_b = P_{R^{-1}\tilde b}=P_{\tilde b}}$. These stipulations hold in the well-studied case of $W\sim \mathcal{N}(0,\sigma_W^2)$, $W\sim \mathcal{N}(0,\sigma_b^2)$ when $R\in SO(D)$, due to the invariance of the $b_j b_j$ in $P_b={\rm exp}(-b_jb_j/2\sigma_b^2)$, and similarly for Gaussian $P_W$.

The result holds more generally, for any invariant $P_W$ and $P_b$, which as we will discuss in Section \ref{sec:learning} includes the case of correlated parameters, as is relevant for learning.

\paragraph{Example: $SO(\din)$ Input Symmetry.}

We now demonstrate an example of neural networks with density invariant under $SO(\din)$ input rotations, provided that the input layer parameters are drawn from an invariant distribution. 

We will take a linear input layer and turn off the bias for simplicity, since it may be trivially included as in the $SO(\dout)$ output symmetry above.  The network function is $f_i(x) = g_{ij}(W_{jk} x_k)$, $W\sim P_W$, and the input rotation $R\in SO(\din)$ acts as  $x_i \mapsto x_i' = R_{ij} x_j$. The output of the input layer is the preactivation for the rest of the network $g$, which has parameters $\theta_g$. The transformed correlators are 
\begin{align} 
  G'^{(n)}_{i_1\dots i_n}&(x_1',\dots,x_n') = \mathbb{E}[f_{i_1}(R_{k_1l_1}x^1_{l_1})\dots f_{i_n}(R_{k_nl_n}x^n_{l_n})] \nonumber \\
  &= \frac{1}{Z_\theta} \int DW \,  D\theta_g \, g_{i_1j_1}(W_{j_1k_1}R_{k_1l_1}x^1_{l_1})\dots g_{i_nj_n}(W_{j_nk_n}R_{k_nl_n}x^n_{l_n}) P_W P_{\theta_g} \nonumber \\
  &= \frac{1}{Z_\theta} \int |R^{-1}|  D\tilde W \,  D\theta_g \, g_{i_1j_1}(\tilde W_{j_1l_1}x^1_{l_1})\dots g_{i_nj_n}(\tilde W_{j_nl_n}x^n_{l_n}) P_{R^{-1} \tilde W}  P_{\theta_g} \nonumber \\
  &= \mathbb{E}[f_{i_1}(x_1)\dots f_{i_n}(x_n)] = G^{(n)} (x_1,\dots,x_n),
\end{align}
where we have changed the $x$ subscript label to a superscript to make room for indices, and again the important second-to-last equality holds when $P_W$ is invariant under $R\in SO(\dout)$. This again holds for $W_{ij}\sim \mathcal{N}(0,\sigma_W^2)$, but also for any distribution $P_W$ constructed from $SO(\dout)$ invariants. See \cite{Cohen_2021} for $SO(\din)$ input symmetry of the NNGP kernel.

\paragraph{$SU(\dout)$ Output Symmetry} 
We also demonstrate that a linear complex-valued output layer, given in details in Appendix \eqref{app:su}, leads to $SU(\dout)$ invariant networks densities provided that last linear layer weight and bias distributions are invariant. For clarity we leave off the bias term; it may be added trivially similar to Eqn. \eqref{eq-SO-output}. This network is defined by $\mathbf{f}_{i} = \mathbf{W}_{ij}g_j(x, \theta_g)$, and transforms as $\mathbf{f}_{i} \mapsto S_{ij}\mathbf{f}_{j},~ \mathbf{f}^{\dagger}_{k} \mapsto \mathbf{f}^{\dagger}_{l}S^{\dagger}_{lk}$ under an invertible matrix transformation by $SU(\dout)$ group element $S$. A necessary condition for symmetry is that the only non-zero correlation functions have an equal number of $f$ and $f^\dagger$'s, as in \eqref{app:complexnpt2}, which transform as
\begin{align}
&G^{'(2n)}_{i_1 \cdots i_{2n}} (x'_{1}, \cdots, x'_{2n}) = \mathbb{E} \left[ S_{i_1j_1} \mathbf{f}_{j_1}(x_{p_1})  \cdots S_{i_n j_n} \mathbf{f}_{j_n}(x_{p_n}) \mathbf{f}^{\dagger}_{j_n}(x_{p_{n+1}})S^{\dagger}_{j_{n+1} i_{n+1}}  \cdots  \mathbf{f}^{\dagger}_{j_{2n}}(x_{p_{2n}})  S^{\dagger}_{j_{2n} i_{2n}} \right] \nonumber \\
&=   \frac{1}{4Z_\theta}  \int\, D\mathbf{W}D\mathbf{W}^{\dagger}D\theta_g S_{i_1j_1} \, \big(\mathbf{W}_{j_1 k_1}g_{k_1}(x_{p_1}, \theta_g)\big) \cdots S_{i_n j_n} \big(\mathbf{W}_{j_n k_n} g_{k_n}(x_{p_n}, \theta_g) \big)  \nonumber \\ 
&\big( g^{\dagger}_{k_{n+ 1}}(x_{p_{n+ 1}}, \theta_g) \mathbf{W}^{\dagger}_{k_{n+1} j_{n+1}} \big) S^{\dagger}_{j_{n+1} i_{n+1}} \cdots  \big( g^{\dagger}_{k_{2n}}(x_{p_{2n}}, \theta_g) \mathbf{W}^{\dagger}_{k_{2n} j_{2n}} \big) S^{\dagger}_{j_{2n} i_{2n}}  P_{\mathbf{W}, \mathbf{W}^{\dagger}} P_{\theta_g} \nonumber \\
&=   \frac{1}{4Z_\theta}  \int\, |(S^{\dagger})^{-1}||S^{-1}| D\mathbf{\tilde{W}}D\mathbf{\tilde{W}}^{\dagger}D\theta_g  \big(\mathbf{\tilde{W}}_{i_1 k_1}g_{k_1}(x_{p_1}, \theta_g)\big) \cdots\big( \mathbf{\tilde{W}}_{i_n k_n}g_{k_n}(x_{p_n}, \theta_g)\big) \nonumber \\
& \big( g^{\dagger}_{k_{n+1}}(x_{p_{n+1}}, \theta_g)\mathbf{\tilde{W}}^{\dagger}_{k_{n+1} i_{n+1}}\big) \cdots\big( g^{\dagger}_{k_{2n}}(x_{p_{2n}}, \theta_g)\mathbf{\tilde{W}}^{\dagger}_{k_{2n} i_{2n}}\big) P_{S^{-1}\mathbf{\tilde{W}}  , \mathbf{\tilde{W}}^{\dagger} {S^{\dagger}}^{-1} }  P_{\theta_g}  \nonumber \\
&= \mathbb{E} \left[ \mathbf{f}_{i_1}(x_{p_1})  \cdots \mathbf{f}_{i_n}(x_{p_n}) \mathbf{f}^{\dagger}_{i_{n+1}}(x_{p_{n+1}})\cdots  \mathbf{f}^{\dagger}_{i_{2n}}(x_{p_{2n}})  \right]= G^{(2n)}_{i_1 \cdots i_{2n}} (x_{1}, \cdots, x_{2n}),
\end{align}
where $\{p_1, \cdots , p_{2n} \}$ is any permutation of $\{1, \cdots, 2n \}$ as in \eqref{app:su}, and the crucial second-to-last equality holds when $|S|^{-1}=1$, and $P_{\mathbf{W}, \mathbf{W}^{\dagger}} = P_{\mathbf{\tilde{W}}  , \mathbf{\tilde{W}}^{\dagger}  }$. This stipulation holds true, for example, when $S \in SU(\dout) $, and $\text{Re}(\mathbf{W}), \text{Im}(\mathbf{W}) \sim \mathcal{N}(0, \sigma^2_W)$, due to the $SU(\dout)$ invariance of $\text{Tr}(\mathbf{W}^{\dagger} \mathbf{W})$ in $P_{\mathbf{W}  , \mathbf{W^{\dagger}}  } = \exp( - \text{Tr}(\mathbf{W}^{\dagger} \mathbf{W})/2\sigma^2_W ) $. For more details, please see appendix \eqref{app:su}.

\paragraph{Example: Translation Input Symmetry and T-layers.} We now study architectures with translation ($T$) invariant network densities, which arise when the correlation functions are invariant under input translations $x \mapsto x + c$, $\forall c\in \mathbb{R}^\din$, the usual notion of translations in physics. Our translations are more general than the pixel translations often studied in computer vision; for instance, in one dimension a pixel shift to the right is induced by the choice $c_i = -x_i + x_{(i+1)\%d}$.

To arrive at $T$-invariant network densities via correlation functions, we first define the \emph{$T$-layer}, a standard linear layer with deterministic weight matrix $W$ and uniform bias on the circle, $b_i\sim \mathcal{U}(S^{1})$, where we also map the $Wx$ term to the circle by taking it mod 1 (written as \% 1) before adding the already $S^{1}$-valued bias. Since such a layer is defined by the hyperparameter weight matrix $W$, we label it $T_W$, with parameters $b$. Suppressing obvious indices, we write the $T$-layer as $T_W(x) = ( Wx \, \% \, \, 1 ) +  b $. Under translations of the input to the $T$-layer, we have $T_W(x+c) = ( Wx) \, \% \, \, 1 + (Wc) \, \% \, \, 1  + b =: ( Wx )\, \% \, \, 1 + b'$, where $b'$ and $b$ are equally probable bias draws and the periodic boundary condition of the $S^1$ solves problems that arise in the presence of boundaries of a uniform distribution on a subset of $\mathbb{R}$. 

The $T$-layer may be prepended to any neural network to arrive at a new one with $T$-invariant density. 
Consider any neural network $g_{\varphi}:\mathbb{R}^N \to \bR^\dout$, to which we prepend $T_W: \bR^\din \to \bR^N$ to form a new network $f_\theta(x) = g_\varphi(T_W(x))$,
where $\theta = \{\varphi,b\}$. The transformed correlation functions are translation invariant,
\begin{equation}
  G^{(n)}_{i_1,\dots,i_n}(x_1+c,\dots, x_n+c) = G^{(n)}_{i_1,\dots,i_n}(x_1,\dots, x_n) \qquad \forall n,
\end{equation}
which follows by absorbing the shift via $b' = (Wc \, \% \, \, 1 ) + b$, using $db' = db$, and renaming variables. See appendix \eqref{app:translation} for details. Thus, the density $P_f$ is translation invariant.

The $T$-layer is compatible with training since it is differentiable everywhere except when the layer input is $\equiv 0 \text{ mod } 1$, i.e. an integer. When doing gradient descent, the mod operation is treated as the identity, which gives the correct gradient at all non-integer inputs to the layer, and thus training performs well on typical real-world datasets. 

\section{Symmetry-via-Duality and Deep Learning \label{sec:learning}}
We now various aspects of relating symmetry, deduced via duality, and learning.

\paragraph{Preserving Symmetry During Training, via Duality.} It may sometimes be useful to preserve a symmetry (deduced via duality) during training that is present in the network density at initialization. Corollary \ref{cor} allows symmetry-via-duality to be utilized at any time, relying on the invariance properties of $P_{\theta_T}$ (and therefore $P_\theta$). The initialization symmetry is preserved if the invariance properties of $P_\theta$ that ensured symmetry at $t=0$ persist at all times.

While this interesting matter deserves a systematic study of its own, here we study it in the simple case of continuous time gradient descent. The parameter update $d\theta_i/dt = -\partial \mathcal{L}/\partial{\theta_i}$, where $\mathcal{L}$ is the loss function, induces a flow in $P_\theta$ governed by 
\begin{align}
    \frac{\partial P_\theta(t)}{\partial t} =  \displaystyle \left(\frac{\partial}{\partial\theta_i} \frac{\partial}{\partial\theta_i}\mathcal{L}\right)\, P_\theta(t)+ \frac{\partial P_\theta(t) }{\partial \theta_i} \frac{\partial \mathcal{L}}{\partial \theta_i}, \label{eqn:pdfdynamics}
\end{align}
the update equation for $P_\theta(t)$. If $P_\theta(t)$ is invariant at initialization ($t=0$), then the update is invariant provided that $\partial^2 \mathcal{L}/(\partial \theta_i)^2$ is invariant and the second term is invariant.

When these conditions are satisfied, the symmetry of the network initialization density is preserved throughout training. However, they must be checked on a case-by-case basis. As a simple example, consider again the $SO(D)$ output symmetry from Section \ref{sec:symmetryviaduality}. Absorbing the action on output into parameters as before, the $(\partial/\partial{\theta_i})(\partial/\partial{\theta_i})$ is itself invariant, and therefore the first term in \eqref{eqn:pdfdynamics} is invariant when $\mathcal{L}$ is invariant. If additionally
\begin{equation}
  \frac{\partial P_\theta(t) }{\partial \theta_i} = I_P\, \theta_i \qquad \qquad \frac{\partial \mathcal{L}}{\partial \theta_i} = I_\mathcal{L}\, \theta_i
\end{equation}
for $\theta$-dependent invariants $I_P$ and $I_\mathcal{L}$, then the second term is invariant as well, yielding an invariant symmetry-preserving update. See Appendix \eqref{app:symtrain} for a detailed example realizing these conditions.

\paragraph{Supervised Learning, Symmetry Breaking, and the One-point Function.}
A priori, it is natural to expect that symmetry breaking helps training: a non-zero one-point function or mean $\smash{G^{(1)}_i(x) = \bE[f_i(x)]}$ of a trained network density is crucial to making non-zero predictions in supervised learning, and if a network density at initialization exhibits symmetry, developing a one-point function during supervised learning usually breaks it. Contrary to this intuition, we will see in experiments that symmetry in the network density at initialization can help training.

To develop these ideas and prepare for experiments,
we frame the discussion in terms of an architecture with symmetry properties that are easily determined via duality: a 
network with a linear no-bias output layer from an $N$-dimensional hidden layer to $D$-dimensional network output. The network function is $\smash{f_i(x) = W^{l}_{ij}g_j(x)}$, $i=1,\dots,\dout$, where $g_j(x)$ is the post-activation of the last hidden layer, with parameters $\theta_g$.  Output weights in the final ($l^{th}$) layer are initialized as 
\begin{equation}
  \label{eq:outputweightbreaking}
  W^{l}_{ij} \sim \mathcal{N}(\mu_{W^{l}},1/\sqrt{N}),\,\,\,\,  \forall i < k+1 \qquad \qquad 
  W^{l}_{ij} \sim \mathcal{N}(0,1/\sqrt{N}),\,\,\,\, \forall i \geq k + 1,
\end{equation}
where $k$ is a hyperparameter that will be varied in the experiments. By a simple extension of our $SO(\dout)$ output symmetry result, this network density has $SO(\dout-k)$ symmetry. The symmetry breaking is measured by the one-point function  
\begin{equation}
G^{(1)}_i(x) =\bE[f_i(x)] =  N \mu_{W^{l}}\,\,\bE_{\theta_g}[g_i(x)]\neq 0,\ \,\,\,\, \forall i < k,
\end{equation}
and zero otherwise. Here, $\neq 0$ means as a function; for some $x$, $G_i^{(1)}(x)$ may evaluate to $0$. This non-zero mean breaks symmetry in the first $k$ network components, and since $SO(\dout)$ symmetry is restored in the $\mu_{W^{l}}\to 0$ limit, $\mu_{W^{l}}$ and $k$ together control the amount of symmetry breaking.

\paragraph{Symmetry and Correlated Parameters.}  Learning-induced flows in neural network densities lead to correlations
between parameters, breaking any independence that might exist in the parameter priors. Since symmetry-via-duality relies only on an invariant $P_\theta$, it can apply in the case of correlations, when $P_\theta$ does not factorize.
In fact, this is the generic case: a density $P_\theta$ which is symmetric due to being constructed from group invariants is not, in general, factorizable.  For instance, the multivariate Gaussian $\smash{P_\theta = \text{exp}(- x_i x_i/ 2\sigma^2)}$ with $i=1,\dots, m$ is constructed from the $SO(m)$ invariant $x_i x_i$ and leads to independent parameters due to factorization. However, provided a density normalization condition is satisfied, additional $SO(l)$-invariant terms $c_n \, (x_i x_i)^n$ (or any other Casimir invariant) may be added to the exponent which preserve symmetry, but break independence for $n>1$.

As an example, consider again the case with $SO(\dout)$ output
symmetry with network function $f(x) = L(g(x,\theta_g))$ where $L:\bR^N \to \bR^\dout$ is a linear layer, but now draw its weights and biases from symmetric parameter distributions with quartic non-Gaussianities, 
\begin{equation}
W_{ij} \sim P_W = e^{-\frac{{\rm Tr} (W^T W) }{2\sigma_{W}^2} - \lambda\, ({\rm Tr}\, (W^T W) )^2} \qquad b_i \sim P_b = e^{-\frac{b\cdot b}{2\sigma_{b}^2} - \lambda\, (b \cdot b)^2}.
\end{equation}
By construction these distributions are invariant under $SO(\dout)$, and therefore the function-space density is as well. However, independence is broken for $\lambda \neq 0$. Training could also mix in parameters from other layers, yielding a non-trivial joint distribution which is nevertheless invariant provided that the final layer parameter-dependence arises only through $SO(D)$ invariants.

Such independence-breaking networks provide another perspective on neural networks and GPs. Since the NNGP correspondence relies crucially on the central limit theorem, and therefore independence of an infinite number of parameters as $N\to \infty$, we may break the GP to a non-Gaussian process not only by taking finite-$N$, but also by breaking independence, as with $\lambda \neq 0$ above. In this example, symmetry-via-duality requires neither the asymptotic $N\to \infty$ limit nor the independence limit. 

Independence breaking introduces potentially interesting non-Gaussianities into function-space densities, which likely admit an effective field theory (EFT) akin to the finite-$N$ EFT treatment developed in \cite{Halverson_2021}. We leave this treatment for future work.

\paragraph{Symmetry and the Neural Tangent Kernel.} Gradient descent training of a neural network $f_\theta$ is governed by the Neural Tangent Kernel (NTK) \cite{Jacot2018NeuralTK}, $\smash{\hat \Theta(x,x') = \partial_{\theta_i} f_\theta(x) \, \partial_{\theta_i} f_\theta(x')}$. Since $\hat \Theta(x,x')$ depends on concrete parameters associated to a fixed neural network draw, it is not invariant.
However, the NTK converges in appropriate large-$N$ limits to a kernel $\Theta$ that is deterministic, due to the appearance of ensemble averages, allowing for the study of symmetries of $\Theta(x,x')$ via duality.

As an example, consider a neural network with a linear output layer (mapping from $\smash{\bR^N\to \bR^\dout})$, $\smash{f_i(x) = W^{l}_{ij}g_j(x)}/\sqrt{N}$, $W^l \sim \mathcal{N}(0,\sigma_{W}^2)$, where $g_j(x)$ is the post-activation of the last hidden layer and we have turned off bias for clarity; it may be added trivially. The corresponding NTK is 
\begin{equation}
  \label{eq:ntkinv}
\hat \Theta_{i_1i_2}(x,x') = \frac{1}{N} \left( g_{j}(x, \theta_{g}) \, g_{j}(x', \theta_{g}) \, \delta_{i_1 i_2} + W^{l}_{i_1 j_1} W^{l}_{i_2 j_2}  \frac{\partial g_{j_1}(x, \theta_{g})}{\partial \theta^{k}_{g}} \frac{\partial g_{j_2}(x', \theta_{g})}{\partial \theta_{g}^{k}}  \right).
\end{equation}
This depends on the concrete draw $f$ and is not invariant.
However, as $N \to \infty$, $\hat \Theta$  at initialization becomes the deterministic NTK 
\begin{equation}
\Theta_{i_1i_2}(x,x') = \delta_{i_1i_2} \, \mathbb{E}\big[g^j(x, \theta_g)g^j(x', \theta_g) \big] \nonumber 
+ \mathbb{E}[W^l_{i_1j_1}W^l_{i_2j_2}] \mathbb{E}\Big[ \frac{\partial g_{j_1}(x, \theta_g)}{\partial \theta^k_g}\frac{\partial g_{j_2}(x', \theta_g)}{\partial \theta^k_g}\Big].
\end{equation}
The transformation $f_i \to R_{ij} f_j$ for $R\in SO(D)$ acts only on the first factor in each term, which themselves may be shown to be invariant, rendering $\Theta_{i_1i_2}(x,x')$ invariant under $SO(D)$. Alternatively, a related calculation shows that $\mathbb{E}[\Theta_{i_1i_2}(x,x')]$ is also $SO(D)$ invariant. 

Such results are more general, arising similarly in other architectures according to Corollary \ref{cor}. Though the deterministic NTK at $t=0$ is crucial in the linearized regime, invariance of $\Theta$ may also hold during training, as may be studied in examples via invariance of $P_\theta$.
See \cite{Cohen_2021} for $SO(D)$ input symmetry of the deterministic NTK $\Theta$.

\section{Experiments} \label{sec:experiments}

We carry out two classes of experiments\footnote{We provide an implementation of our code at \url{https://github.com/keeganstoner/nn-symmetry}.} testing symmetry-via-duality. In the first, we demonstrate that the amount of symmetry in the network density at initialization affects training accuracy.  In the second, presented in Appendix \ref{app:so-expts}, we demonstrate how symmetry may be tested via numerically computed correlators.

\paragraph{Does Symmetry Affect Training?}
\begin{figure}[t]
  \includegraphics[scale=.5]{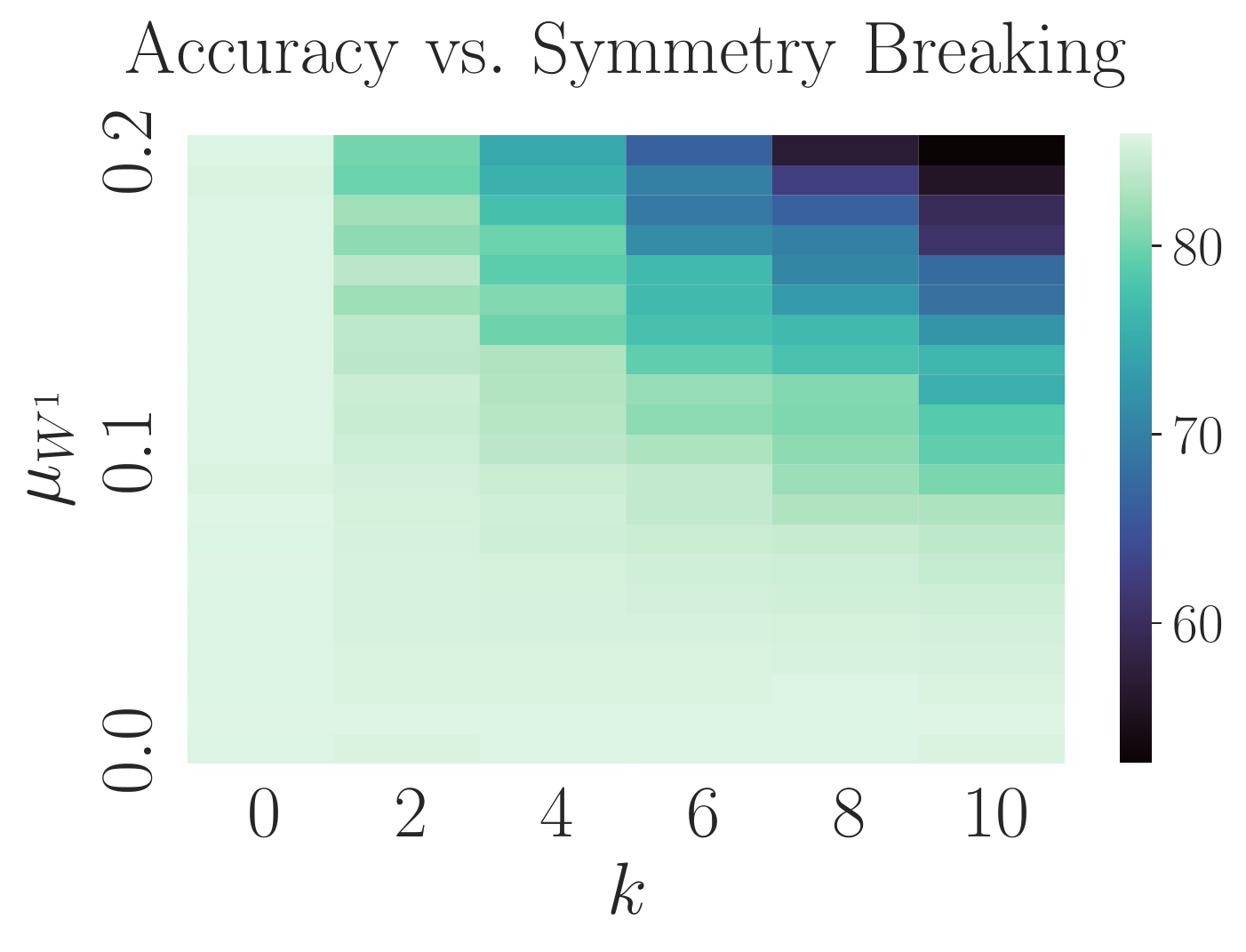}
  \includegraphics[scale=.5]{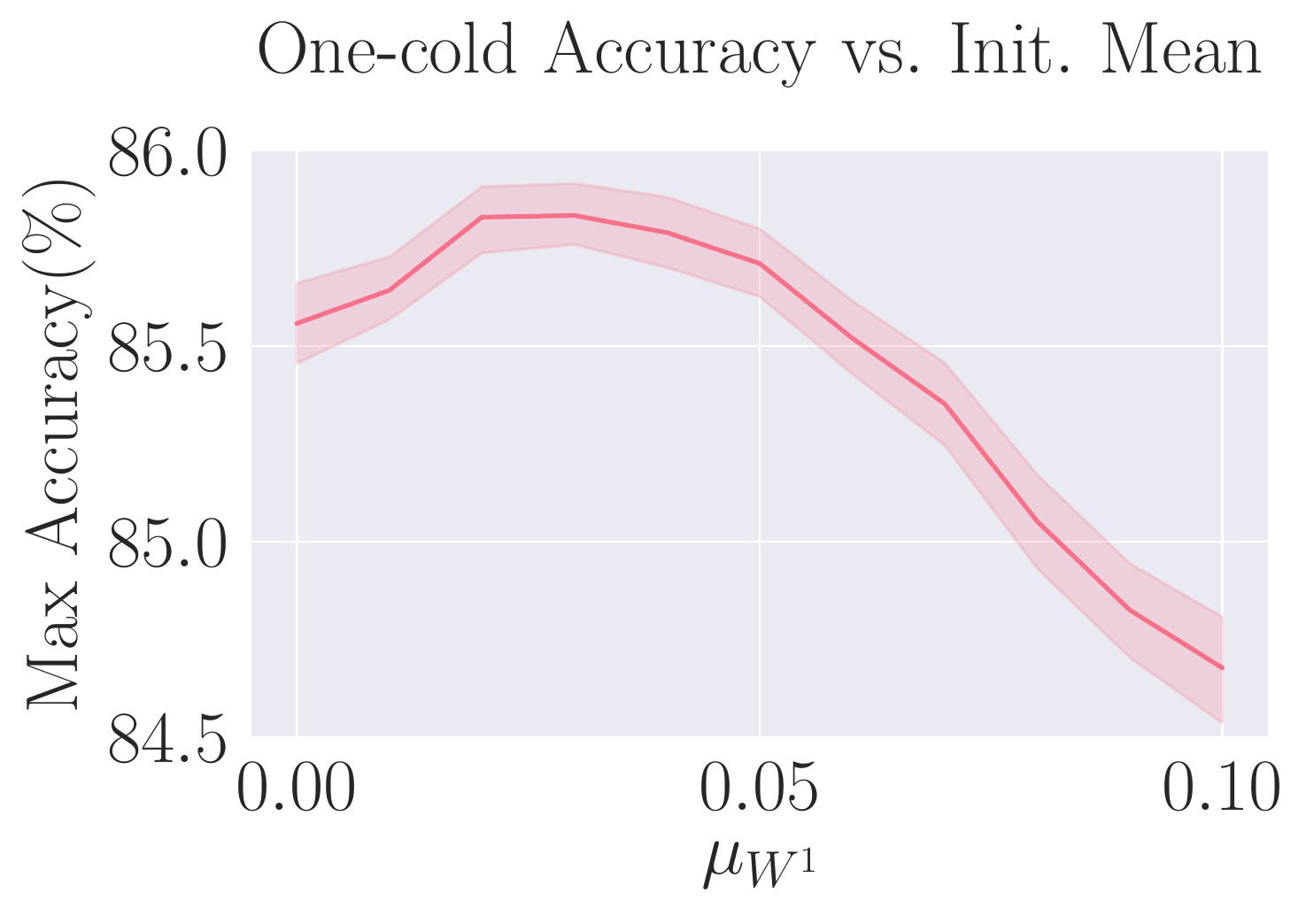}
    \caption{Test accuracy \%age on Fashion-MNIST. \emph{(Left):} Dependence on symmetry breaking parameters $\mu_W$ and $k$ for one-hot encoded labels. The error is presented in Appendix \eqref{app:exp-details}. \emph{(Right):} Dependence on $\mu_W$ for one-cold encoded labels, showing the $95\%$ confidence interval. }
  \label{fig}
  \end{figure}

We now wish to test the ideas from Section \ref{sec:learning} on how the amount of symmetry at initialization affects test accuracy after training, as controlled by the hyperparameters $\mu_{W^{1}}$ and $k$; see \cite{kuningangulisymmetrybreaking} for another analysis of symmetry breaking and learning.
We further specify the networks discussed there by choosing a single-layer network ($l = 1$, $\din = 1$) with ReLU non-linearities (i.e., $g_j$ is the post-activation of a linear layer), $N=50$, weights of the first linear layer 
$W^{0}\sim \mathcal{N}(0, 1/\sqrt{\din})$, and weights of the output initialized as in \eqref{eq:outputweightbreaking}.
All networks were trained on the Fashion-MNIST dataset \cite{DBLP:journals/corr/abs-1708-07747} for $20$ epochs with MSE loss, with one-hot (or one-cold) encoded class labels, leading to network outputs with dimension $\dout = 10$, and therefore symmetry $SO(10-k)$ at initialization.

In the first experiment, we study how the amount of rotational symmetry breaking at initialization affects test accuracy on Fashion-MNIST with one-hot encoded class labels. We vary the amount of symmetry breaking by taking $k\in\{0,2,4,6,10\}$ and $\mu_{W^1} \in \{0.0,\dots,0.2\}$ with $.01$ increment. Each experiment is repeated $20$ times with learning rate $\eta = .001$. For each $(k,\mu_{W^1})$ pair, the mean of the maximum test accuracy across all $20$ experiments is plotted in Figure \ref{fig} (LHS).
We see that performance is highest for networks initialized with $\mu_{W^{1}} = 0$ or $k=0$, i.e. with an $SO(D)$ symmetric initialization density, and decreases significantly with increasing amounts of symmetry breaking (increasing $\mu_{W^{1}}$ and $k$), contrary to the intuition discussed in Section \ref{sec:learning}. See Appendix \eqref{app:exp-details} for more details about the experiments. 

If supervised learning breaks symmetry via developing a non-trivial one-point function (mean), but we see that symmetry at initialization helps training in this experiment, then what concept is missing? 

It is that symmetry breaking at initialization could be in the wrong direction, i.e. the initialization mean is quite different from the desired trained mean, which (if well-trained) approximate ground truth labels. In our experiment, the initialization mean is 
\begin{equation}
  G^{(1)}_i(x) =\bE[f_i(x)] =  50 \mu_{W^{1}}\,\,\bE_{W^0}[\,{\rm ReLU}(W^0_{ij} x_j)\,]\neq 0,
\end{equation}
which will in general be non-zero along all output components. It is "in the wrong direction" since class labels are one-hot encoded and therefore have precisely one non-zero entry. Furthermore, even for means in the right direction, the magnitude could be significantly off.

In the second experiment, we test this idea by imposing a better match between the initialization mean and the class labels. To do so, we keep the network fixed and instead one-cold encode the class labels, i.e. instead of class $i$ being encoded by the unit vector $e_i \in \bR^D$, it is encoded by $\mathbb{1}-e_i$, where $\mathbb{1}$ is the vector of ones.
At initialization 
we find that $\smash{\bE_{W^0}[\,{\rm ReLU}(W^0_{ij} x_j)\,]\sim 0.5 - 1.0}$ for Fashion-MNIST, and therefore $G^{(1)}_i(x) \sim 25 \mu_{W^{1}} - 50 \mu_{W^{1}}$ is of the right order of magnitude to match the ones in the one-cold vectors for $\mu_{W^{1}}\sim .02 - .04$.
Relative to the first experiment, this experiment differs only in the one-cold encoded class labels and the fixed value $k=10$, which ensures complete symmetry breaking in the prior for $\mu_{W^1} \neq 0$.

 We see from Figure \ref{fig} (\textit{right}) that performance improves until $\mu_{W^{1}} \sim .02 - .04$, but then monotonically decreases for larger $\mu_{W^{1}}$. By construction, the symmetry breaking is much closer to the correct direction than in the first experiment, but the magnitude of the initialization means affects performance: the closer they are to the $D-1$ ones in the one-cold encoding, the better the performance. The latter occurs for $\mu_{W^{1}}\sim .02 - .04$ according to our calculation, which matches the experimental result.

\section{Conclusion \label{sec:discussion}}

We introduce symmetry-via-duality, a mechanism that allows for the determination of symmetries of neural network functional densities $P_f$, even when the density is unknown.  The mechanism relies crucially two facts: i) that symmetries of a statistical system may also be determined via their correlation functions; and ii) that the correlators may be computed in parameter space. The utility of parameter space in determining symmetries of the network density is a hallmark of duality in physical systems, in this case, \duality duality. 

We demonstrated that invariance of correlation functions ensures the invariance of $\textit{Df}\, P_f$, which yields the invariance of the density $P_f$ itself in the absence of a Green-Schwarz mechanism. Symmetries were categorized into input and output symmetries, the analogs of spatial and internal symmetries in physics, and a number of examples of symmetries were presented, including $SO(D)$ and $SU(D)$ symmetries at both input and output. In all calculations, the symmetry transformation induces a transformation on the input or output that may be absorbed into a transformation of network parameters $\theta_T$, and invariance of the correlation functions follows from invariance of $D\theta_T P_{\theta_T}$. The invariance of $D\theta P_\theta$ also follows, since $\theta_T$ are by definition the only parameters that transform.

The mechanism may also be applied at any point during training, since it relies on the invariance of $P_\theta$. If duality is used to ensure the symmetry of the network density at initialization, then the persistence of this symmetry during training requires that $P_\theta$ remains symmetric at all times. Under continuous time gradient descent, the flow equation for $P_\theta$ yields conditions preserving the symmetry of $P_\theta$. We also demonstrated that symmetry could be partially broken in the initialization density, that symmetry-via-duality may also apply in the case of non-independent parameters, and that the Neural Tangent Kernel may be invariant under symmetry transformations.

Our analysis allows for different amounts of symmetry in the network density at initialization, leading to increasing constraints on the density with increasing symmetry. Accordingly, it is natural to ask whether this affects training. To this end,
we performed Fashion-MNIST experiments with different amounts of network density symmetry at initialization. The experiments demonstrate that symmetry breaking helps training when the associated mean is in the direction of the class labels, and entries are of the same order of magnitude. However, if symmetry is broken in the wrong direction or with too large a magnitude, performance is worse than for networks with symmetric initialization density.

\newpage

\section*{Acknowledgements}
We thank Sergei Gukov, Joonho Kim, Neil Lawrence, Magnus Rattray, and Matt Schwartz for discussions. We are especially indebted to S{\'e}bastian Racani{\`e}re, Danilo Rezende, and Fabian Ruehle for comments on the manuscript.
J.H. is supported by NSF CAREER grant PHY-1848089. This work is supported by the National Science Foundation under Cooperative Agreement PHY-2019786 (The NSF AI Institute for Artificial Intelligence and Fundamental Interactions).

\bibliography{refs}
\bibliographystyle{unsrt}

\clearpage
\newpage

\onecolumn
\appendix

{\Large \textbf{Supplementary Material}}

\section{Proofs and derivations}
\subsection{Symmetry from Correlation Functions}

We begin by demonstrating the invariance of network correlation functions under transformations
that leave the functional measure and density invariant.
 Consider a transformation 
  \begin{equation}
    f'(x) =\Phi(f(x'))
  \end{equation}
that leaves the functional density invariant, i.e.
\begin{equation}
\label{app:fdinv}
  D[\Phi f]\, e^{-S[\Phi f]} = \textit{Df}\,\, e^{-S[f]}.
\end{equation}
Then we have 
\begin{align}
\bE[f(x_1)\dots f(x_n)] &=  \frac{1}{Z_f}\int \textit{Df}\,\, e^{-S[f]}\, f(x_1)\dots f(x_n) \\
&=  \frac{1}{Z_f}\int \textit{Df}\,'\, e^{-S[f']}\, f'(x_1)\dots f'(x_n) \\
&=  \frac{1}{Z_f}\int D[\Phi f]\, e^{-S[\Phi f]}\, \Phi(f(x_1'))\dots \Phi(f(x_n')) \\
&=  \frac{1}{Z_f}\int \textit{Df}\,\, e^{-S[f]}\, \Phi(f(x_1'))\dots \Phi(f(x_n')), \\
&= \bE[\Phi(f(x_1'))\dots \Phi(f(x_n'))]
\end{align}
where the second to last equality holds due to the invariance of the functional
density.  This completes the proof of \eqref{eqn:correlatorconstraint}; See, e.g., \cite{DiFrancesco:1997nk} for a QFT analogy.

For completeness we wish to derive the same result for infinitesimal output transformations, where
the parameters of the transformation depend on the neural network input; in physics language, these are
called infinitesimal gauge transformations.

The NN output, transformed by an infinitesimal parameter $\omega_a(x)$, is $f'(x) = \Phi(f(x'))= f(x') + \delta_{\omega}f(x')$, where $ \delta_{\omega}f(x') = - \iota \omega_a(x') T_a f(x')$; $T_a$ is the generator of the transformation group. Corresponding output space log-likelihood transforms as $S \mapsto S - \int\, dx' \, \partial_\mu \, j^\mu_a(x')\, \omega_a(x')$, for a current $j^\mu_a(x')$ that may be computed. The transformed $n$-pt function at $O(\omega)$ is given by,
\begin{equation}
\begin{split}
&\mathbb{E}[ f'(x_1)\cdots f'(x_n) ] = \frac{1}{Z} \int \textit{Df}\,' \, \Phi\big(f(x'_1)\cdots f(x'_n) \big) e^{ - S} \nonumber \\
& = \frac{1}{Z}\int \textit{Df} \, \Big(\big[f(x'_1)\cdots f(x'_n)\big] + \delta_\omega \big[f(x'_1)\cdots f(x'_n) \big]  \Big)e^{ - S - \int \,dx' \partial_\mu j^\mu_a(x') \omega_a(x')} \\
&= \mathbb{E}[ f(x'_1)\cdots f(x'_n) ] +   \mathbb{E} \big[  \delta_{\omega } [ f(x'_1)\cdots f(x'_n) ] \big] -\int dx' \,  \partial_\mu \mathbb{E}[\, j^\mu_a (x')\, f(x'_1)\cdots f(x'_n)] \, \omega_a(x'), \label{eq:wardid1}
\end{split}
\end{equation}
where we obtain second and last equalities under the assumption that functional density is invariant, following \eqref{app:fdinv}, and invariance of function-space measure, $\textit{Df}\,' = \textit{Df}\,$, respectively. 

Following the $\omega$-independence of L.H.S. of \eqref{eq:wardid1}, $O(\omega)$ terms on R.H.S. must cancel each other, i.e. 
\begin{equation}
\label{eq:wardid11}
\int dx'\,\Big[ \partial_\mu\, \mathbb{E} \big[\,  j^\mu_a \, f(x'_1)\cdots f(x'_n) \big]  +  \mathbb{E}\Big[ \sum_{i= 1}^n  f(x'_1)\cdots T_a f(x'_i) \cdots  f(x'_n)  \Big] \delta(x'-x'_i)\Big]\omega(x') = 0,
\end{equation}
for any infinitesimal function $\omega(x')$. Thus, the coefficient of $\omega(x')$ in above integrand vanishes at all $x'$, and we have the following by divergence theorem
\begin{align}
-  \iota \int dx'\, \mathbb{E}\Big[ \sum_{i= 1}^n  f(x'_1)\cdots T_a f(x'_i) \cdots  f(x'_n)  \Big] \delta(x'-x'_i) &=  \int_{\Sigma} ds_{\mu} \mathbb{E} \big[ j^\mu_a \, f(x'_1)\cdots f(x'_n) \big]. \label{eq:wardid2}
\end{align}
Taking hypersurface ${\Sigma}$ to infinity does not affect the integral in \eqref{eq:wardid2}, therefore in $\lim_{R_{\Sigma} \rightarrow \infty}$, if $\mathbb{E} \big[ j^\mu_a f(x'_1)\cdots f(x'_n) \big]$ dies sufficiently fast, we obtain 
\begin{eqnarray}
\iota \sum_{i= 1}^n \mathbb{E} \big[ f(x'_1)\cdots T_a f(x'_i) \cdots  f(x'_n) \big] = 0  =\delta_\omega\, \mathbb{E} \big[\,  f(x'_1)\cdots f(x'_n)  \big], \label{eq:wardid3}
\end{eqnarray}
a statement of invariance of correlation functions under infinitesimal input-dependent transformations. 

Thus, we obtain the following invariance under finite / infinitesimal, input-dependent/independent transformations, whenever $\textit{Df}\, = \textit{Df}\,'$,
\begin{eqnarray}
\mathbb{E} \big[ f'(x_1)\cdots f'(x_n) \big] = \mathbb{E} \big[ f(x'_1)\cdots f(x'_n) \big] . \label{eq:wardid4}
\end{eqnarray}
\eqref{eq:wardid4} is same as \eqref{eqn:correlatorconstraint}, completing the proof.

\subsection{$SU(\dout)$ Output Symmetry}
\label{app:su}
We show the detailed construction of $SU(\dout)$ invariant network densities, for networks with a complex linear output layer, when weight and bias distributions are $SU(\dout)$ invariant. 

The network is defined by $f(x) = L(g(x, \theta_g))$, for a final affine transformation $L$ on last postactivation $g(x, \theta)$; $x$ and $\theta_g$ are the inputs and parameters until the final linear layer, respectively. As $SU(\dout)$ is the rotation group over complex numbers, $SU(\dout)$ invariant NN densities require complex-valued outputs, and this requires complex weights and biases in layer $L$. Denoting the real and imaginary parts of complex weight $\mathbf{W}$ and bias $\mathbf{b}$ in layer $L$ as $W^1, W^2, b^1, b^2$ respectively, we obtain $\mathbf{W}, \mathbf{b}$ distributions as $P_{\mathbf{W}, \mathbf{W}^{\dagger}} = P_{W^1}P_{W^2}$ and $P_{\mathbf{b}, \mathbf{b}^{\dagger}} = P_{b^1}P_{b^2}$. The simplest $SU(\dout)$ invariant structure is 
\begin{equation}
\text{Tr}[\mathbf{W}^\dagger \mathbf{W}] = \mathbf{W}^{\ast}_{\alpha \beta}  \mathbf{W}_{\alpha \beta} = W^{1}_{\alpha \beta} W^{1}_{\alpha \beta} + W^{2}_{\alpha \beta} W^{2}_{\alpha \beta} = \text{Tr}({W^{1}}^T W^{1}) + \text{Tr}({W^{2}}^T W^{2})  , \label{eq:consistency0}
\end{equation}
and similarly for bias. To obtain an $SU$-invariant structure in $P_{\mathbf{W}, \mathbf{W}^{\dagger}}$ as a sum of $SO$-invariant structures from products of $P_{W^1}, P_{W^2}$, all three PDFs need to be exponential functions, with equal coefficients in $P_{W^1},P_{W^2}$. Therefore, starting with $SO(\dout)$-invariant real and imaginary parts $W^1, W^2 \sim \mathcal{N}(0, \sigma^2_W)$ and $b^1, b^2 \sim \mathcal{N}(0, \sigma^2_b)$, one can obtain the simplest $SU(\dout)$ invariant complex weight and bias distributions, given by $P_{\mathbf{W}, \mathbf{W}^{\dagger}} = \exp( - \text{Tr}[\mathbf{W}^\dagger \mathbf{W}]/2\sigma^2_W )$, $P_{\mathbf{b}, \mathbf{b}^{\dagger}} = \exp( - \text{Tr}[\mathbf{b}^\dagger \mathbf{b}]/2\sigma^2_b )$ respectively. 

We want to express the network density and its correlation functions entirely in terms of complex-valued outputs, weights and biases, therefore, we need to transform the measures of $W^1,W^2,b^1,b^2$ into measures over $\mathbf{W}, \mathbf{b}$. As $DW^1DW^2 = |J|D\mathbf{W}D\mathbf{W}^{\dagger}$, for Jacobian of $\begin{bmatrix}
W^{1} \\ W^{2}
\end{bmatrix} =  \begin{bmatrix}
 \frac12 & \frac12 \\ \frac{\iota}{2} & - \frac{\iota}{2}
\end{bmatrix} \begin{bmatrix}
\mathbf{W} \\ \mathbf{W}^{\dagger}
\end{bmatrix} $, we obtain $DW^1DW^2Db^1Db^2 = |J|^2 D\mathbf{W}D\mathbf{W}^{\dagger}D\mathbf{b}D\mathbf{b}^{\dagger}$, and $|J|^2 = 1/4$. With this, the $n$-pt function for any number of $f$'s and $f^\dagger$'s becomes the following,
\begin{align}
\label{app:complexnpt1}
&G^{(n)}_{i_1, \cdots, i_n }(x_1, \cdots, x_n) =  \mathbb{E}[f_{i_1}(x_1)\cdots f_{i_r}(x_r)\,f^\dagger_{i_{r+1}}(x_{r+1}) \cdots f^\dagger_{i_n}(x_n) ] \nonumber \\
&= \frac{1}{4Z_\theta}  \int  D\mathbf{W}D\mathbf{W}^{\dagger}D\mathbf{b} D\mathbf{b}^{\dagger}D\theta_g \Big[ \mathbf{W}_{i_1j_1} g_{j_1}(x_1, \theta_g) + \mathbf{b}_{i_1}  \Big] \cdots 
\Big[ \mathbf{W}_{i_rj_r} g_{j_r}(x_r, \theta_g) + \mathbf{b}_{i_r}  \Big] \nonumber \\
&\Big[ g^{\dagger}_{j_{r+1}}(x_{r+1}, \theta_g)\mathbf{W}^{\dagger}_{j_{r+1}i_{r+1}}  + \mathbf{b}^{\dagger}_{i_{r+1}} \Big] \cdots \Big[  g^{\dagger}_{j_n}(x_n, \theta_g)\mathbf{W}^{\dagger}_{j_ni_n} + \mathbf{b}^{\dagger}_{i_n} \Big] e^{ - \frac{\text{Tr}(\mathbf{W}^\dagger \mathbf{W})}{2\sigma_{W}^{2}}    - \frac{\text{Tr}( \mathbf{b}^\dagger \mathbf{b}) }{2\sigma_{b}^{2}}} P_{\theta_g},
\end{align}
where $Z_\theta$ is the normalization factor. We emphasize that the transformation of $f_i$ and $f^\dagger_j$ only transforms two indices inside the trace in $\text{Tr}(\mathbf{W}^\dagger \mathbf{W})$; it is invariant in this case, and also when all four indices transform. 

From the structure of $P_{\mathbf{W}, \mathbf{W}^{\dagger}}$ and $P_{\mathbf{b}, \mathbf{b}^{\dagger}}$, only those terms in the integrand of \eqref{app:complexnpt1}, that are functions of $W^\dagger_{i_s} W_{i_s}$ and $b^\dagger_{i_t} b_{i_t}$ alone, and not in product with any number of $W_{i_u}, W^{\dagger}_{i_u}, b_{i_u}, b^{\dagger}_{i_u}$ individually, result in a non-zero integral. 
Thus, we have the only non-vanishing correlation functions from an equal number of $f$'s and $f^\dagger$'s. We hereby redefine the correlation functions of this complex-valued network as
\begin{align}
\label{app:complexnpt2}
G^{(2n)}_{i_1,\cdots, i_{2n}}(x_1, \cdots , x_{2n}) :=\mathbb{E}[f_{i_1}(x_{p_1})\cdots f_{i_n}(x_{p_n})\, f^{\dagger}_{i_{n+1}}(x_{p_{n+1}})\cdots f^{\dagger}_{i_{2n}}(x_{p_{2n}}) ] ,
\end{align}
where $\{p_1, \cdots , p_{2n} \}$ can be any permutation of set $\{1, \cdots, 2n \}$.

\subsection{$SU(\din)$ Input Symmetry}
We now show an example of neural networks densities invariant under $SU(\din)$ input transformations, provided that input layer parameters are drawn from an $SU(\din)$ invariant distribution.  

We will take a linear input layer and turn off bias for simplicity, as it may be trivially included as in $SU(\dout)$ output symmetry. $SU(\din)$ group acts on complex numbers, therefore network inputs and input layer parameters need to be complex, such a network function is $f_i = g_{ij}(\mathbf{W}_{jk}x_k)$. The distribution of $\mathbf{W}$ is obtained from products of distributions of its real and imaginary parts $W^1, W^2$. Following $SU(\dout)$ output symmetry demonstration, the simplest $SU(\din)$ invariant $P_{\mathbf{W}, \mathbf{W}^{\dagger}}$ is obtained when $W^1, W^2 \sim \mathcal{N}(0,\sigma^2_W)$ are both $SO(\din)$ invariant, we get $P_{\mathbf{W}, \mathbf{W}^{\dagger}} = \exp( - \text{Tr}[\mathbf{W}^\dagger \mathbf{W}]/2\sigma^2_W )$. The measure of $\mathbf{W}$ is obtained from the measures over $W^1,W^2$ as $DW^1DW^2 = |J|D\mathbf{W}D\mathbf{W}^{\dagger}$, with $|J|=1/2$. Following a similar analysis as \eqref{app:su}, the only non-trivial correlation functions are
\begin{align}
 &   G^{(2n)}_{i_1,\cdots, i_{2n}}(x_1, \cdots , x_{2n}) :=\mathbb{E}[f_{i_1}( x_{p_1}) \cdots f_{i_n}(x_{p_n})\, f^{\dagger}_{i_{n+1}}(x^{\dagger}_{p_{n+1}}  )\cdots f^{\dagger}_{i_{2n}}(x^{\dagger}_{p_{2n}} ) ]  \nonumber \\
& = \frac{1}{2Z_\theta}  \int D\mathbf{W}D\mathbf{W}^{\dagger}D\theta_g\, g_{i_1j_1} (\mathbf{W}_{j_1k_1}x^{p_1}_{k_1})  \cdots g_{i_nj_n} (\mathbf{W}_{j_nk_n}x^{p_n}_{k_n}) \, (x^{\dagger}{}^{p_{n+1}}_{k_{n+1}} \mathbf{W}^{\dagger}_{k_{n+1}j_{n+1}}) g^{\dagger}_{j_{n+1}i_{n+1}} \nonumber \\
&  \cdots (x^{\dagger}{}^{p_{2n}}_{k_{2n}} \mathbf{W}^{\dagger}_{k_{2n}j_{2n}}) g^{\dagger}_{j_{2n}i_{2n}} e^{ - \frac{\text{Tr}(\mathbf{W}^\dagger \mathbf{W})}{2\sigma_{W}^{2}}} P_{\theta_g},
\end{align}
where $\{p_1, \cdots , p_{2n} \}$ is any permutation over $\{1, \cdots, 2n \}$, and we have changed the $x$ subscript label to a superscript to make room for the indices. 

Under input rotations $x_i \mapsto S_{ij}x_j,~ x^{\dagger}_k \mapsto x^{\dagger}_l S^{\dagger}_{lk}$ by $S \in SU(\din)$, the correlation functions transform into 
\begin{align}
G'^{(2n)}_{i_1, \cdots, i_{2n} }&(x'_{1}, \cdots, x'_{2n}) = \mathbb{E}[f_{i_1}(S_{k_1l_1} x^{p_1}_{l_1}) \cdots f_{i_n}(S_{k_nl_n} x^{p_n}_{l_n})\, f^{\dagger}_{i_{n+1}}(x^{\dagger}{}^{p_{n+1}}_{l_{n+1}}S^{\dagger}_{l_{n+1}k_{n+1}}  )\cdots f^{\dagger}_{i_{2n}}(x^{\dagger}{}^{p_{2n}}_{l_{2n}}S^{\dagger}_{l_{2n}k_{2n}}  ) ] \nonumber \\
     =& \,  \frac{1}{2Z_\theta}  \int\, D\mathbf{W}D\mathbf{W}^{\dagger}D\theta_g \,g_{i_1j_1}(\mathbf{W}_{j_1 k_1}S_{k_1l_1}x^{p_1}_{l_1})  \cdots g_{i_nj_n}(\mathbf{W}_{j_n k_n}S_{k_nl_n}x^{p_n}_{l_n}) \nonumber \\
&  (x^{\dagger}{}^{p_{n+1}}_{l_{n+1}} S^{\dagger}_{l_{n+1}k_{n+1}} \mathbf{W}^{\dagger}_{k_{n+1}j_{n+1}})   g^{\dagger}_{j_{n+1}i_{n+1}} \cdots (x^{\dagger}{}^{p_{2n}}_{l_{2n}} S^{\dagger}_{l_{2n}k_{2n}} \mathbf{W}^{\dagger}_{k_{2n}j_{2n}}) g^{\dagger}_{j_{2n}i_{2n}}    P_{\mathbf{W}, \mathbf{W}^{\dagger}} P_{\theta_g} \nonumber \\
=&\,   \frac{1}{2Z_\theta}  \int\, |{S^{\dagger}}^{-1}| |S^{-1}| D\mathbf{\tilde{W}}D\mathbf{\tilde{W}}^{\dagger}D\theta_g  \,g_{i_1j_1}(\mathbf{\tilde{W}}_{j_1 l_1}x^{p_1}_{l_1})  \cdots g_{i_nj_n}(\mathbf{\tilde{W}}_{j_n l_n}x^{p_n}_{l_n}) \nonumber \\
& (x^{\dagger}{}^{p_{n+1}}_{l_{n+1}} \mathbf{\tilde{W}}^{\dagger}_{l_{n+1}j_{n+1}} )g^{\dagger}_{j_{n+1}i_{n+1}} \cdots (x^{\dagger}{}^{p_{2n}}_{l_{2n}} \mathbf{\tilde{W}}^{\dagger}_{l_{2n}j_{2n}} )g^{\dagger}_{j_{2n}i_{2n}} P_{S^{-1}\mathbf{\tilde{W}}  , \mathbf{\tilde{W}}^{\dagger}S^{\dagger}{}^{-1}  } P_{\theta_g}  \nonumber \\
=&\, \mathbb{E}[f_{i_1}(x_{p_1}) \cdots f_{i_n}( x_{p_n})\, f^{\dagger}_{i_{n+1}}(x^{\dagger}_{p_{n+1}} )\cdots f^{\dagger}_{i_{2n}}(x^{\dagger}_{p_{2n}} ) ] = G^{(2n)}_{i_1, \cdots, i_{2n} }(x_{1}, \cdots, x_{2n}).
\end{align}
The crucial second-to-last equality holds for $|(S^{\dagger})^{-1}| |S^{-1}|=1$, as is the case here; further we need $P_{\mathbf{W}, \mathbf{W}^{\dagger}} = P_{\mathbf{\tilde{W}}  , \mathbf{\tilde{W}}^{\dagger}  }$, this stipulation holds true when $\text{Re}(\mathbf{W}), \text{Im}(\mathbf{W}) \sim \mathcal{N}(0, \sigma^2_W)$, due to the $SU$-invariance of $\text{Tr}(\mathbf{W}^{\dagger} \mathbf{W})$.

\subsection{Translation Input Symmetry} \label{app:translation}
We demonstrate an example of network densities that remain invariant under continuous translations on input space, when the input layer weight is deterministic and input layer bias is sampled from a uniform distribution on the circle, $b \sim \mathcal{U}(S^{1})$. We will map the weight term to the circle by taking it mod 1, (i.e. \, \% 1). 

The network output $f_i(x)= g_{ij}( ( W_{jk}x_k )\, \% \, \, 1 + b_j )$ transforms into $f'(x') = g_{ij} ( (W_{jk}x_k )\, \% \, \, 1 +  b'_j ) $ under translations of inputs $x_k \mapsto x_k + c_k$, where $b'_j = (W_{jk}c_k) \, \% \, \, 1 + b_j$. With a deterministic $W$, the network parameters are given by $\theta = \{\phi, b \}$, and $Db' = Db$. The transformed $n$-pt function is
\begin{align}
   & G'^{(n)}_{i_1, \cdots, i_n }(x_1, \cdots, x_n)= \mathbb{E}[f'_{i_1}(x'_n)\cdots f'_{i_n}(x'_n)] \nonumber \\
&= \frac{1}{Z_\theta}  \int DbD\phi \, g_{i_1j_1}(( W_{j_1k_1}x_{k_1} + W_{j_1k_1}c_{k_1})\, \% \, \, 1 + b_{j_1} )  \cdots \nonumber \\ & \qquad \qquad \qquad \cdots g_{i_nj_n}( ( W_{j_nk_n}x_{k_n} + W_{j_nk_n}c_{k_n} )\, \% \, \, 1 + b_{j_n} )  P_b P_{\phi} \nonumber \\
&= \frac{1}{Z_\theta}  \int Db' D\phi \, g_{i_1j_1}( ( W_{j_1k_1}x_{k_1} )\, \% \, \, 1 + b'_{j_1}  )  \cdots g_{i_nj_n}( ( W_{j_nk_n}x_{k_n} )\, \% \, \, 1  + b'_{j_n} )  P_{b'} P_{\phi} \nonumber \\
& = \mathbb{E}[f_{i_1}(x_n)\cdots f_{i_n}(x_n)] = G^{(n)}_{i_1, \cdots, i_n }(x_1, \cdots, x_n),
\end{align}
where the crucial third-to-last equality holds when $P_b = P_{b'}$. This stipulation is true as $\mathbb{E}[b^k] = \mathbb{E}[b’^k]$ for any $k$ when $b \sim \mathcal{U}(S^{1})$ and $W$ is deterministic; since the layer is valued on the circle with circumference 1, we know that any bias value is equally probable. Thus $b$ and $b’$ have identical moment generating functions and $P_b = P_{b’}$.

\subsection{$Sp(\dout)$ Output Symmetry}
We also demonstrate an example of network densities that remain invariant under the compact symplectic group $Sp(\dout)$ transformations on output space. 

The compact symplectic $Sp(\dout)$ is the rotation group of quaternions, just as $SU$ is the rotation group of complex numbers. Thus, a network with linear output layer would remain invariant under compact symplectic group, if last linear layer weights and biases are quaternionic numbers, drawn from $Sp(\dout)$ invariant distributions. We define the network output $f_i(x) = W_{ij}g_{j} (x, \theta_g)+ b_i$ as before, with parameters $W_{ab} = W_{ab,0} + i W_{ab,1} + j W_{ab, 2} + k W_{ab, 3}$ and $b_{a} = b_{a,0} + i b_{a,1} + j b_{a,2} + k b_{a, 3}$, such that Hermitian norms $\text{Tr}(W^{\dagger}W) = W^{\dagger}_{ab}W_{ab} = \sum_{i=0}^{3}W^2_{ab,i}$ and $\text{Tr}(b^{\dagger}b) = b^{\dagger}_{a}b_{a} = \sum_{i=0}^{3}b^2_{a,i}$ are compact symplectic $Sp(\dout)$ invariant by definition, where the conjugate of a quarternion $q = a + i b + j c + k d$ is $q^*= a - ib -jc -kd$. The distributions of $W, b$ are obtained as products of distributions of the components $W_0,W_1,W_2,W_3$ and $b_0,b_1,b_2,b_3$ respectively. Following the $SU(\dout)$ symmetry construction, we can obtain the simplest $Sp(\dout)$ invariant $P_{W,W^{\dagger}}$ and $P_{b,b^{\dagger}}$ when these are functions of the Hermitian norm, and PDF of each component $P_{W_i}$ is an exponential function of $SO(\dout)$ invariant term $\text{Tr}(W^T_iW_i)$ with equal coefficient, similarly with bias. Starting with $W_0,W_1,W_2,W_3 \sim \mathcal{N}(0, \sigma^2_W)$, and $b_0,b_1,b_2,b_3 \sim \mathcal{N}(0, \sigma^2_b)$, we get $Sp(\dout)$ invariant quaternionic parameter distributions $P_{W,W^{\dagger}} = \exp( - \text{Tr}(W^{\dagger}W)/ 2 \sigma^2_W)$ and $P_{b,b^{\dagger}} = \exp( - \text{Tr}(b^{\dagger}b)/ 2 \sigma^2_b)$. We also obtain the measures over $W,b$ from measures over $W_i$ and $b_i$, e.g. $DWDW^{\dagger} = |J|DW_0DW_1DW_2DW_3$. 
Following an analysis similar to \eqref{app:su}, it can be shown that the only non-trivial correlation functions of this quaternionic-valued network are 
\begin{align}
    {G}^{(n)}_{i_1, \cdots, i_{2n}} (x_{1}, \cdots, x_{2n}) =& \frac{|J|^2}{Z_\theta} \int DWDW^\dagger Db Db^\dagger f_{i_1}(x_{p_1}) \cdots f_{i_n}(x_{p_n})\,f^{\dagger}_{i_{n+1}}(x_{p_{n+1}}) \cdots \nonumber \\
   & ~ f^{\dagger}_{i_{2n}}(x_{p_{2n}})e^{ - \frac{\text{Tr}({W}^{\dagger} {W})}{2\sigma_{W}^{2}}  - \frac{\text{Tr}({b}^{\dagger} {b})}{2\sigma_{b}^{2}}} P_{\theta_g},
\end{align}
for $\{p_1, \cdots , p_{2n} \}$ any permutation over $\{1, \cdots, 2n \}$. Under $Sp(\dout)$ transformation of outputs $f_i \mapsto S_{ij}f_j,~ f^{\dagger}_k \mapsto f^{\dagger}_{l}S^{\dagger}_{lk}$, by $S \in Sp(\dout)$ in quaternionic basis, the correlation functions transform as
\begin{align}
G'^{(2n)}_{i_1 \cdots i_{2n}} &(x'_{1}, \cdots, x'_{2n}) = \mathbb{E} \left[ S_{i_1j_1} f_{j_1}(x_{p_1})  \cdots S_{i_n j_n} f_{j_n}(x_{p_n})\, f^{\dagger}_{j_{n+1}}(x_{p_{n+1}})S^{\dagger}_{j_{n+1}i_{n+1}}\cdots f^{\dagger}_{j_{2n}}(x_{p_{2n}})S^{\dagger}_{j_{2n}i_{2n}}  \right] \nonumber \\
=&\,   \frac{|J|^2}{Z_\theta}  \int\, DWDW^{\dagger}Db Db^{\dagger}D\theta_g S_{i_1j_1} \big(W_{j_1 k_1}g_{k_1}(x_{p_1}, \theta_g) + b_{j_1} \big) \cdots S_{i_n j_n} \big(W_{j_n k_n} g_{k_n}(x_{p_n}, \theta_g) \nonumber \\
&+ b_{j_n} \big)  \big(g^{\dagger}_{k_{n+1}}(x_{p_{n+1}}, \theta_g )W^{\dagger}_{k_{n+1}j_{n+1}} +  b^{\dagger}_{j_{n+1}}\big)S^{\dagger}_{j_{n+1}i_{n+1}}  \cdots \big(g^{\dagger}_{k_{2n}}(x_{p_{2n}}, \theta_g )W^{\dagger}_{k_{2n}j_{2n}} +  b^{\dagger}_{j_{2n}}\big)\nonumber \\
&S^{\dagger}_{j_{2n}i_{2n}}     P_{W, W^{\dagger}}P_{b, b^{\dagger}} P_{\theta_g} \nonumber \\
=&\,   \frac{|J|^2}{Z_\theta}  \int\, |S^{-1}||S^{\dagger}{}^{-1}| D\tilde{W}D{\tilde{W}}^{\dagger}D\tilde{b} D{\tilde{b}}^{\dagger}D\theta_g  \big({\tilde{W}}_{i_1 k_1}g_{k_1}(x_{p_1}, \theta_g) + {\tilde{b}}_{i_1} \big) \cdots\big( {\tilde{W}}_{i_n k_n}g_{k_n}(x_{p_n}, \theta_g)    \nonumber \\
&+ {\tilde{b}}_{i_n} \big) \big(g^{\dagger}_{k_{n+1}}(x_{p_{n+1}}, \theta_g ){\tilde{W}}^{\dagger}_{k_{n+1}i_{n+1}} +  {\tilde{b}}^{\dagger}_{i_{n+1}} \big) \cdots  \big(g^{\dagger}_{k_{2n}}(x_{p_{2n}}, \theta_g ){\tilde{W}}^{\dagger}_{k_{2n}i_{2n}} +  {\tilde{b}}^{\dagger}_{i_{2n}} \big)       \nonumber \\
&P_{S^{-1}{\tilde{W}}  ,{\tilde{W}}^{\dagger} S^{\dagger}{}^{-1}  } P_{S^{-1}{\tilde{b}}, {\tilde{b}}^{\dagger}S^{\dagger}{}^{-1} } P_{\theta_g}  \nonumber \\
=&\, \mathbb{E} \left[ {f}_{i_1}(x_{p_1})  \cdots {f}_{i_n}(x_{p_n}) \, {f}^{\dagger}_{i_{n+1}}(x_{p_{n+1}})  \cdots {f}^{\dagger}_{i_{2n}}(x_{p_{2n}})\right] = G^{(2n)}_{i_1 \cdots i_{2n}}(x_{{1}}, \cdots, x_{2n}).
\end{align}
The crucial second-to-last equality holds when $|S^{-1}|=1$, $P_{{W}, {W}^{\dagger}} = P_{{\tilde{W}}  , {\tilde{W}}^{\dagger}  }$, and $P_{{b}, {b}^{\dagger}} =  P_{{\tilde{b}}  , {\tilde{b}}^{\dagger}  }$. These stipulations hold true, for example, when $S \in Sp(\dout) $, $W_0, W_1, W_2, W_3 \sim \mathcal{N}(0, \sigma^2_W)$, and $b_0,b_1,b_2,b_3 \sim \mathcal{N}(0, \sigma^2_b)$, due to the invariance of $\text{Tr}(W^{\dagger} W)$ in $P_{W  , W^{\dagger}  } = \exp( - \text{Tr}(W^{\dagger} W)/2\sigma^2_W ) $, and similarly for $P_{b  , b^{\dagger}  }$. 

\subsection{Preserving Symmetry During Training: Examples} \label{app:symtrain}

We study further the example of an $SO(\dout)$ output symmetry from Section \ref{sec:symmetryviaduality}. Turning off the bias for simplicity, the network function is 
\begin{equation}
  f_i(x) = W_{ij}g_{j} (x),  
\end{equation}
with parameters $\theta = \{W,\theta_g\}$; transformations of $f_i$ may be absorbed into $W$, i.e. $W= \theta_T$.

The network density remains symmetric during training when the updates to $P_\theta$ preserve symmetry; for this example, we showed in Section \ref{sec:learning} that it occurs when $\mathcal{L}$ is invariant and 
\begin{equation}
  \frac{\partial P_\theta }{\partial \theta_i} = I_P\, \theta_i \qquad \qquad \frac{\partial \mathcal{L}}{\partial \theta_i} = I_\mathcal{L}\, \theta_i.
  \label{eq:invconditions}
\end{equation}
Let the initial density be $P_\theta(0) = \exp[ - \sum_{j=1}^{k} a_j (\text{Tr}(\theta^T\theta))^j]$ for $a_j \in \bR$. This clearly satisfies the first condition in \eqref{eq:invconditions}.
An example of $SO$-invariant loss function is 
\begin{align}
  \mathcal{L} & = \sum_{x,y} \big( f_i(x)f_i(x) - y_jy_j \big) \nonumber \\ 
  & = \sum_{x,y} \big( W_{il}g_{l}(x) W_{ik}g_{k}(x) - y_jy_j \big),
\end{align}
where $\partial\mathcal{L}/\partial\theta_g$ is invariant because $\mathcal{L}$ is and $\theta_g$ does not transform. Furthermore,
\begin{equation}
\frac{\partial \mathcal{L}}{\partial{W_{mn}}} = 2\,W_{mk}   \left( \sum_{x,y}   g_k(x) g_n(x) \right)
\end{equation}
which satisfies the second condition in \eqref{eq:invconditions}, since the first index is the one that transforms when $W$ absorbs the transformation of $f_i$.

\section{More General $SO$ Invariant Network Distributions}

We will now give an example of an $SO(\dout)$ invariant non-Gaussian network distribution at infinite width, as parameters of the initialized $SO(\dout)$ invariant GP become correlated through training. Such a network distribution can be obtained up to perturbative corrections to the initialized network distribution, if the extent of parameter correlation is small. 

Training may correlate last layer weights $\theta_{ij}$ of a linear network output $f_i = \theta_{ij}g_j(x, \theta_g)$ initialized with $\theta_{ij} \sim \mathcal{N}(0, \sigma^2_\theta)$, such that at a particular training step, we get $\mathcal{P}_{\theta} = e^{ - \frac{1}{2\sigma^2_\theta}\theta^2_{\alpha \beta}- \lambda_\theta\, \theta_{ab}\theta_{ab}\theta_{cd}\theta_{cd} }$ independent from $P_{\theta_g}$, with small $\lambda_\theta$. Correlation functions of this network distribution can be obtained by perturbative corrections to correlation functions of the network distribution at initialization. For example, the $2$-pt function of the correlated network distribution is given by
\begin{align}
\label{eqn:indebr2pt}
        &G^{(2), \text{NGP}}_{i_1 i_2}(x_1, x_2) =  \frac{\int D\theta D\theta_g \, \theta_{i_1 j_1} \theta_{i_2 j_2} \big[ 1  - \lambda_\theta \, \theta^2_{ab} \theta^2_{cd} \big]  g_{j_1}(x_1, \theta_g) g_{j_2}(x_2, \theta_g) e^{-\frac{1}{2\sigma^2_\theta}\theta^2_{\alpha\beta } } \pthetag  }{\int D\theta D\theta_g \,\big[ 1 - \lambda_\theta\, \theta^2_{ab } \theta^2_{cd} \big] e^{-\frac{1}{2\sigma^2_\theta}\theta^2_{\alpha\beta } } \pthetag  } \nonumber \\
    &=  \Big[ \sum_{i_1j_1}  \mathbb{E}[\theta^2_{i_1j_1}]  - \lambda_\theta \, \big( \sum_{i_1j_1 \neq ab \neq cd} \mathbb{E}[ \theta^2_{i_1j_1} ] \mathbb{E}[\theta^2_{ab}] \mathbb{E}[\theta^2_{cd}] + \sum_{i_1j_1 \neq ab} \big(\mathbb{E}[ \theta^2_{i_1j_1}]
    \mathbb{E}[ \theta^4_{ab}] + 2\,\mathbb{E}[\theta^4_{i_1j_1}] \cdot \nonumber \\
    &\mathbb{E}[ \theta^2_{ab}] \big) +  \mathbb{E}[\theta^6_{i_1j_1}] \big)\Big]\Big[1  - \lambda_\theta \big(\sum_{ab} \mathbb{E}[\theta^4_{ab}] + \sum_{ ab \neq cd}  \mathbb{E}[\theta^2_{ab}]\mathbb{E}[\theta^2_{cd}]   \big) \Big]^{-1} \mathbb{E} \big[g_{j_1}(x_1, \theta_g) \,g_{j_1}(x_2, \theta_g)\big] + O(\lambda^2_\theta) \nonumber \\
    &= G^{(2), \text{GP}}_{i_1 i_2}(x_1, x_2) - \lambda_\theta \sum_{i_1j_1 \neq ab} \Big(\mathbb{E}[\theta^6_{i_1j_1}] - \mathbb{E}[\theta^2_{i_1j_1}]\mathbb{E}[\theta^4_{i_1j_1}] - 2\,(\mathbb{E}[\theta^2_{i_1j_1}])^2 \mathbb{E}[\theta^2_{ab}] + 2\,\mathbb{E}[\theta^4_{i_1j_1}]\mathbb{E}[\theta^2_{ab}] \Big)\cdot \nonumber \\
    & \mathbb{E}\big[g_{j_1}(x_1, \theta_g) g_{j_1}(x_2, \theta_g)\big] + O(\lambda^2_\theta). 
    \end{align}
$\mathbb{E}[\theta^{n}_{ij}]$ is evaluated using Gaussian ${P}_{\theta, \text{GP}}=e^{-\frac{\theta^2_{\alpha \beta}}{2\sigma^2_\theta}}$ of initialized network distribution. $O(\lambda_\theta)$ terms in \eqref{eqn:indebr2pt} scale as both $1/N$ and $1/N^2$, as can be seen after properly normalization of  $\theta$ by $\sigma^2_\theta \mapsto \frac{\sigma^2_\theta}{N}$; this `mixed' $1/N$ scaling results from parameter correlations. (We set bias to $0$ everywhere for simplicity, analysis for nontrivial bias follows similar method.)

\section{$SO(\dout)$ Invariance in Experiments} \label{app:so-expts}

The correlator constraint \eqref{eqn:correlatorconstraint} gives testable necessary conditions for a symmetric density. Consider a single-layer fully-connected network, called Gauss-net due to having a Gaussian GP kernel defined by  $f(x)=W_1(\sigma(W_0x + b_0)) + b_1$, where $W_0 \sim \mathcal{N}(0,\sigma_W^2 / \sqrt{\din})$, $W_1 \sim \mathcal{N}(0,\sigma_W^2 / \sqrt{N})$, and $b_0, b_1 \sim \mathcal{N}(0,\sigma_b^2)$, with activation 
$\sigma(x) = {\rm exp}(W_0 x + b_0)/\sqrt{{\rm exp}(2(\sigma_b^2 + \sigma_W^2/\din))}$. 

To test for $SO(\dout)$ invariance via \eqref{eqn:correlatorconstraint}, we measure the average elementwise change in $n$-pt functions before and after an $SO(\dout)$ transformation. 
To do this we generate $2$-pt and $4$-pt correlators at various $\dout$ for a number of experiments and act on them with $1000$ random group elements of a given $SO(\dout)$ group. Each group element is generated by exponentiating a random linear combination of generators of the corresponding algebra, namely
\begin{eqnarray}
R_b = \exp{\left(\sum_{i=1}^{p} \alpha_i\cdot T^{i} \right)},
\end{eqnarray}
for $b = 1,\cdots ,1000$, $p= \text{dim}(SO(\dout)) = \dout(\dout - 1)/2$, $\alpha_i \sim \mathcal{U}(0,1)$ and $T^{i}$ are generators of $\mathfrak{so}(\dout)$ Lie algebra; i.e. $\dout \times \dout$ skew-symmetric matrices written in a simple basis\footnote{
  The generators are obtained 
  by choosing each of the $\frac{\dout(\dout - 1)}{2}$ independent planes of rotations to have a canonical ordering with index $i$, determined by a $(p,q)$-plane.  
  Each $i^{th}$ plane of rotation has a generator matrix $[T^i]_{\dout \times \dout}$ with $T^i_{pq} = -1$, $T^i_{qp} = 1$, for $p<q$, rest $0$. For instance,
  at $\dout=3$, there are $3$ independent planes of rotation formed by direction pairs $\{2,3\}$, $\{1,3\}$ and $\{1,2\}$. For each $i^{\text{th}}$ plane defined by directions $\{p, q\}$, general $SO(3)$ elements $[R^i]_{3 \times 3}$ have $R^i_{pq} = - \sin\theta, R^i_{qp} = \sin \theta, R^i_{qq} = \cos\theta,   R^i_{qq} = \cos\theta, R^i_{rr} = 1$ for $p<q\, , \, r \neq p,q$ and variable $\theta$. Expanding each $R^i$ in Taylor series; the coefficients of $\mathcal{O}(\theta)$ terms are taken to define the generators of $\mathfrak{so}(3)$ as in \eqref{so3}. 
  
}. For example, for $\dout = 3$ we take the standard basis for $\mathfrak{so}(3)$, 

\begin{equation} \label{so3}
  T^1= \begin{bmatrix}
0 & 0 & 0 \\
0 & 0 & -1 \\
0 & 1 & 0
\end{bmatrix},\qquad
T^2 = \begin{bmatrix}
0 & 0 & -1 \\
0 & 0 & 0 \\
1 & 0 & 0
\end{bmatrix} ,\qquad
T^3 = \begin{bmatrix}
0 & -1 & 0 \\
1 & 0 & 0 \\
0 & 0 & 0
\end{bmatrix}
\end{equation}
to generate group elements of $SO(3)$. 

We define the elementwise deviation $\mathcal{M}_n = \text{abs}({G}'^{(n)} - G^{(n)} )$ to capture the change in correlators due to $SO(\dout)$ transformations. Here ${G}'^{(n)}_{i_1, \cdots , i_n}(x_1,\dots,x_n):= R_{i_1 p_1}\cdots R_{i_n p_n} G^{(n)}_{p_1,\cdots, p_n}(x_1,\dots,x_n)$ is the $SO$-transformed $n$-pt function; both $\mathcal{M}_n$ and $G^{(n)}$ have the same rank.

Error bounds for deviation $\mathcal{M}_n$ are determined as $\delta \mathcal{M}_n = \sqrt{( \delta {G}'^{(n)} )^2 + ( \delta G^{(n)} )^2 }$ using the standard error propagation formulae, $\delta G^{(n)}$ equals the average (elementwise) standard deviation of $n$-pt functions across 10 experiments, and $\delta {G}'^{(n)}$ is calculated as the following 
\begin{align}
&\delta {G}'^{(n)} = \frac{1}{\dout^n } \frac{1}{1000}  \sum_{b\,=1}^{1000} \Bigg[  \sum_{t = (i_1,p_1)}^{(i_n,p_n)} \big( R^b_{ i_1 p_1 } \cdots \delta R^b_{t} \cdots  R^b_{i_n p_n} G^{(n)}_{p_1, \cdots, p_n} \big)^2 + \big( R^b_{i_1 p_1}\cdots R^b_{i_n p_n} \delta G^{(n)} \big)^2  \Bigg]^{1/2}  .
 \end{align}
 We have changed the $R$ subscript label $b$ to superscript to make room for the indices. $\delta R_{ij}$ denotes the average error in generated group elements $R$, it is captured by computing $R^TR$ for our experimentally generated matrices $R$ and measuring the magnitude of the off-diagonal elements, which are expected to be zero. We measure an average magnitude of $\mathcal{O}(10^{-18}) =: \delta R$ in these off-diagonal elements. 

\begin{figure}[t]
  \centering
  \hspace{-1.45cm}
\includegraphics[width=0.548\textwidth]{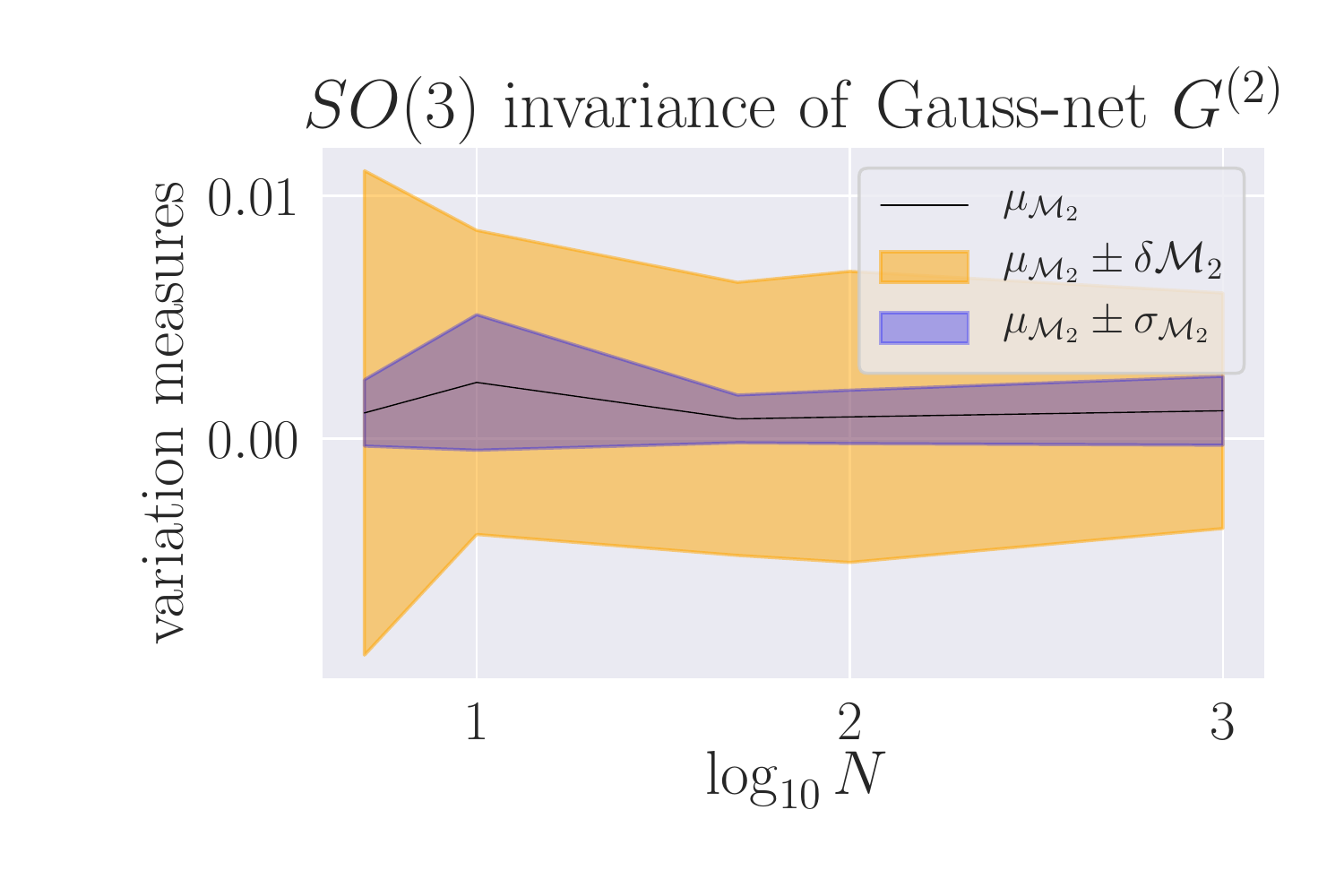}\includegraphics[width=0.548\textwidth]{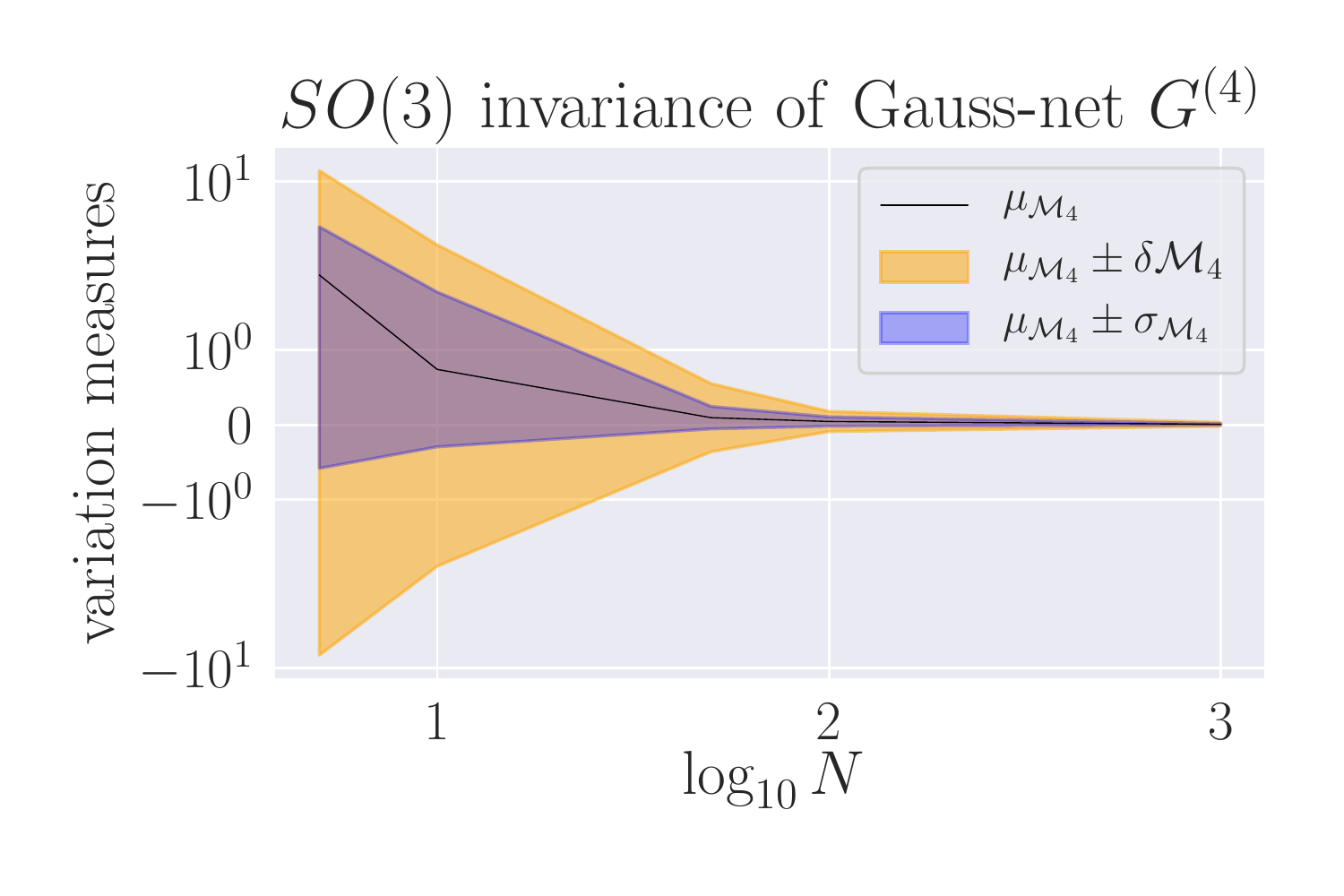} \\   \hspace{-1.45cm}
\includegraphics[width=0.548\textwidth]{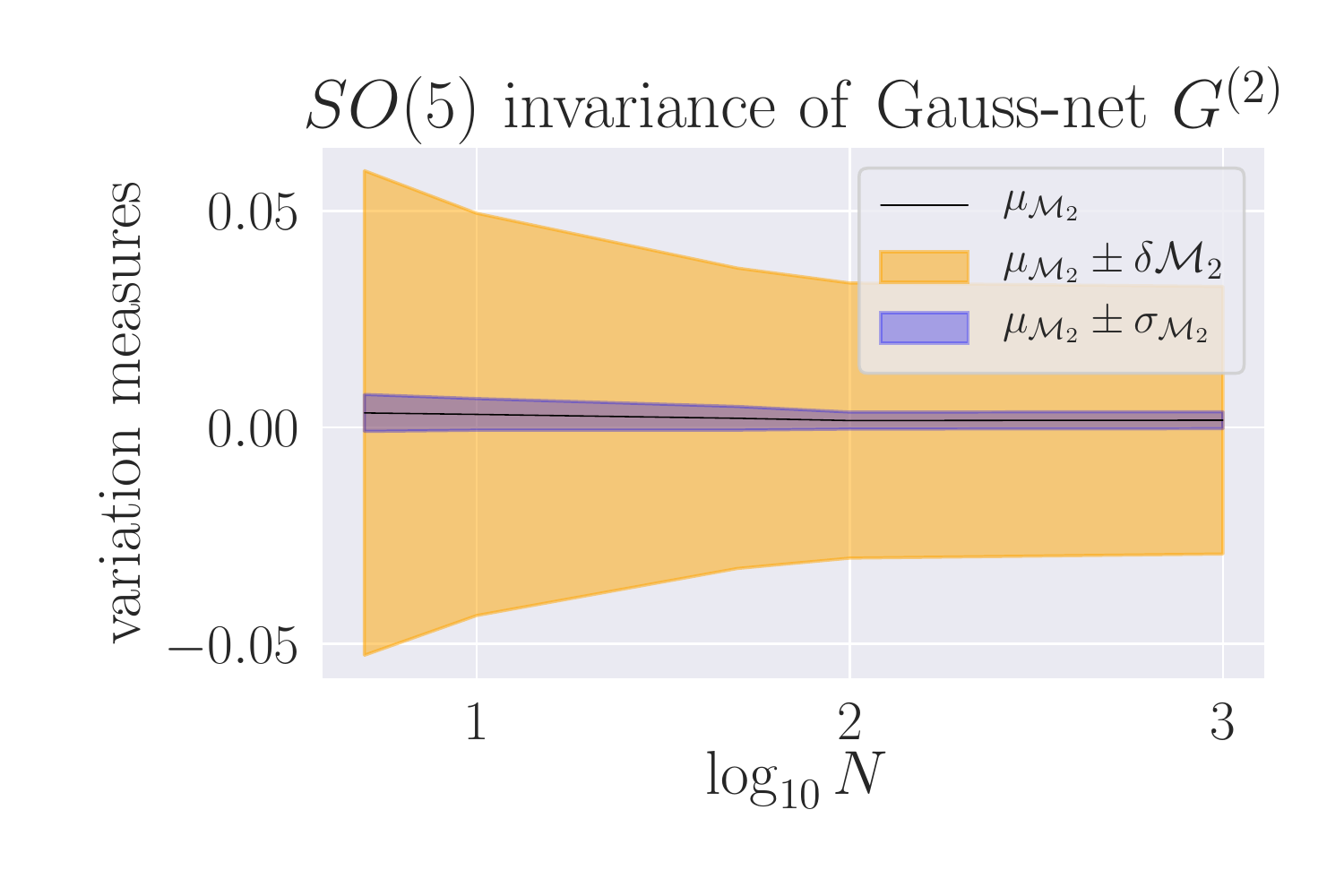}\includegraphics[width=0.548\textwidth]{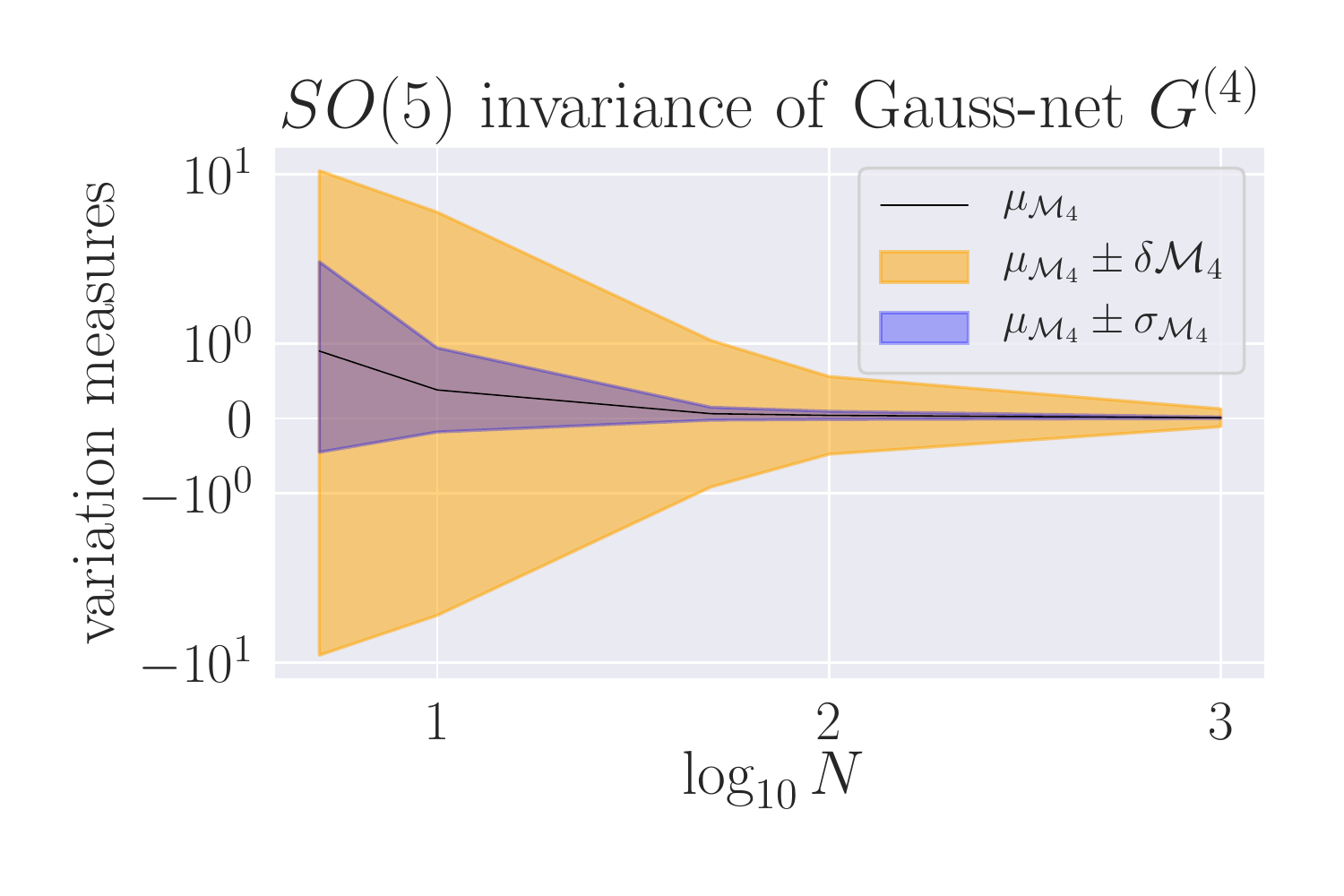}
  \caption{Variation measures of $2$-pt and $4$-pt functions and their predicted error bounds, for $SO(3)$ and $SO(5)$ transformations of $D=3$ and $D=5$ networks, respectively. 
  }
  \label{fig:so5}
\end{figure}

We take the deviation tensors $\mathcal{M}_n$ over $10$ experiments for $SO(3)$ and $SO(5)$ transformations of $2$-pt and $4$-pt functions at $\dout = 3, 5$ respectively, both correlators of each experiment are calculated using $4\cdot 10^6$ and $10^6$ network outputs respectively. An element-wise average and standard deviation across $10$ deviation tensors $\mathcal{M}_n$ are taken and then averaged over, to produce the mean of the $SO$-transformation deviation $\mu_{\mathcal{M}_n}$, and its error $\sigma_{\mathcal{M}_n}$, respectively. We plot $\mu_{\mathcal{M}_n} \pm \sigma_{\mathcal{M}_n}$ (the blue shaded area) in Figure \eqref{fig:so5}, this signal lies well within predicted error bounds of $\pm \delta_{\mathcal{M}_n}$ (in orange), although $\mu_{\mathcal{M}_n}$ deviates significantly from $0$ at low widths, in contradiction to width independence of \eqref{eqn:correlatorconstraint}. This is due to the smaller sample size of parameters in low-width networks, and therefore small fluctuations in the weight and bias draws lead to more significant deviations from the "true" distribution of these parameters. A nonzero mean of the parameters caused by fluctuations leads to a nonzero mean of the function distribution $\langle f \rangle \neq 0$, thus breaking $SO(\dout)$ symmetry. We believe this is a computational artefact and does not contradict $SO$-invariance in \eqref{eqn:correlatorconstraint}.

\section{Experiment Details} \label{app:exp-details}

\begin{figure}[th]
  \centering
\includegraphics[width=0.548\textwidth]{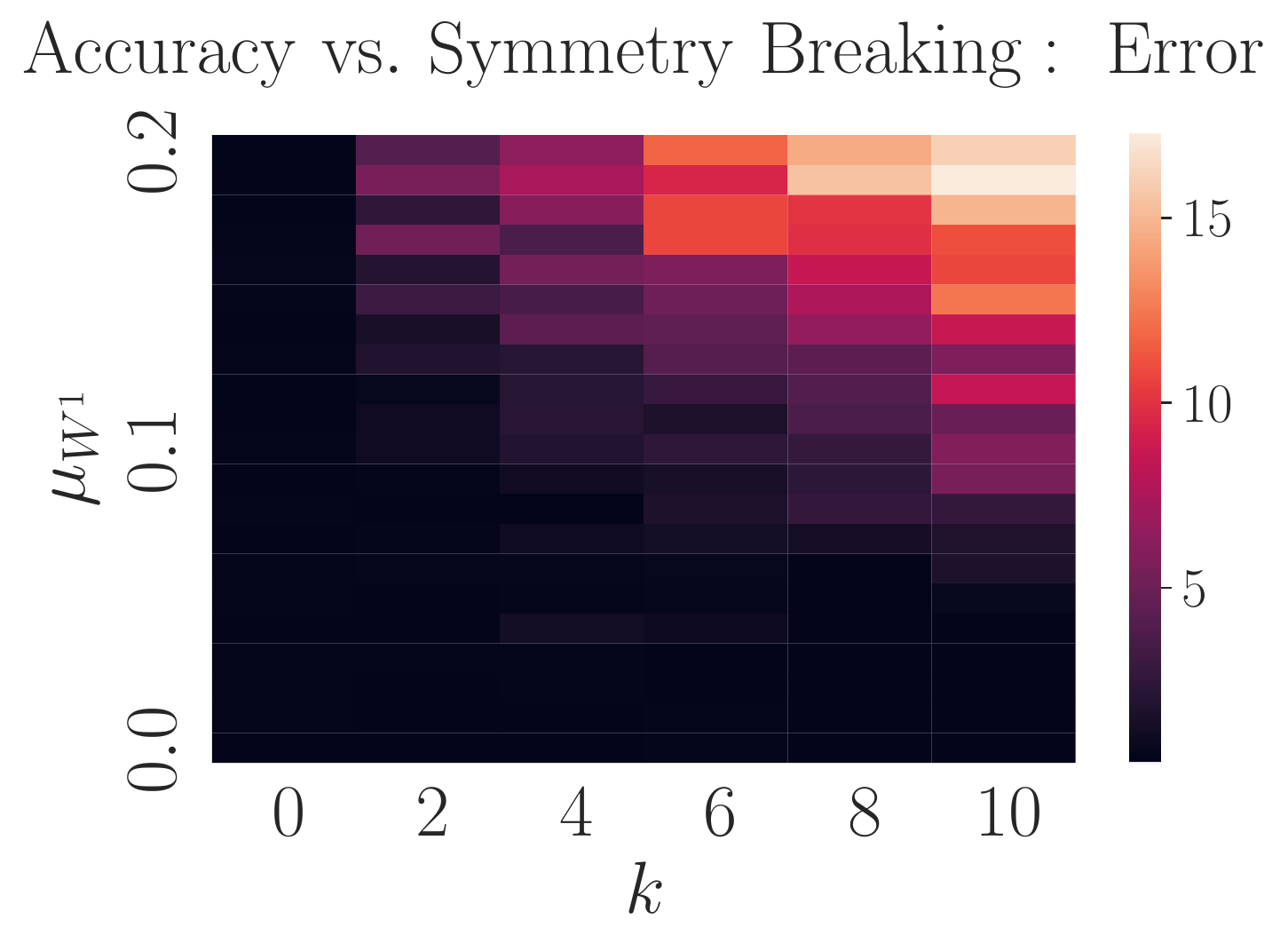}
  \caption{Error in the data from Fig. \eqref{fig}. Color represents the \%age of variation across 20 experiments of the same configuration, computed as the standard deviation normalized by the mean.}
  \label{fig:stdev}
\end{figure}

The experiments in Section \eqref{sec:experiments} were done using Fashion-MNIST under the MIT License\footnote{The MIT License (MIT) Copyright © 2017 Zalando SE, https://tech.zalando.com}, using $60000$ data points for each epoch split with a batch size of $64$, and test batch size of $1000$. Each experiment was run on a 240GB computing node through the Discovery Cluster at Northeastern University and took between 1 and 1.5 hrs to train 20 epochs. The experiments were repeated 20 times for each configuration, and were run with $21 \times 6$ configurations for the left plot in Fig. \eqref{fig}, and $11$ for the right plot. The error in the left plot of Fig. \eqref{fig} is shown in Fig. \eqref{fig:stdev}. 

The experiments in Appendix \eqref{app:so-expts} were done on the same cluster, with the same memory nodes. These took around 24 hours on 10 compute nodes to generate models for each of the $\dout$ values. The $n$-pt functions then took another 6 hours on a single node each.

\section{Comments on Functional Densities} \label{app:densities}
Functional integrals are often treated loosely by physicists: they use them to great effect and experimental agreement in practice, but they are not rigorously defined in general; see, e.g., \cite{Costello2011}.

We follow in this tradition in this work, but would like to make some further comments regarding cases that are well-defined, casting the discussion first into the language of Euclidean QFT, and then bringing it back to machine learning. 

First, the standard Feynman functional path integral for a scalar field $\phi(x)$ is
\begin{equation}
Z = \int D\phi\,\, e^{-S[\phi]},
\end{equation}
but in many cases the action $S[\phi]$ is split into free and interacting pieces
\begin{equation}
S[\phi] = S_F[\phi] + S_{\rm int}[\phi],
\end{equation}
where the free action $S_F[\phi]$ is Gaussian and the interacting action $S_{\rm int}[\phi]$ is non-Gaussian. The free theory, which has $S_{\rm int}[\phi]=0$, is a Gaussian process and is therefore well-defined. When interactions are turned on, i.e. the non-Gaussianities in $S_{\rm int}[\phi]$ are small relative to some scale, physicists compute correlation functions (moments of the functional density) in perturbation theory, truncating the expansion at some order and writing approximate moments of the interacting theory density in terms of a sum of well-defined Gaussian moments, including higher moments.

Second, it is also common to put the theory on a lattice. In such a case the function $\phi: \bR^d \to \bR$ is restricted to a concrete collection of points $\{x_i\}$, $i=1,\dots,m,$ with each $x_i \in \bR^d$. Instead of considering the random function $\phi(x)$ drawn from the difficult-to-define functional density, one instead considers the random variable $\phi(x_i)=: \phi_i$, and the joint distribution on the set of $\phi_i$ defines the lattice theory. One sees this regularly in the Gaussian process literature: evaluated on any discrete set of inputs, the Gaussian process reduces to a standard multivariate Gaussian.

In this work we study functional densities associated to neural networks, and have both the perturbative and especially the lattice understanding in mind when we consider them. In particular, for readers uncomfortable with the lack of precision in defining a functional density, we emphasize that our results can also be understood on a lattice, though input symmetries may be discrete subgroups of those existing in the continuum limit. Furthermore, any concrete ML application involves a finite set of inputs, and for any fixed application in physics or ML one can simply choose the spacing between lattice points to be smaller than the experimental resolution.

\end{document}